\documentclass{WileyMSP-template}
\usepackage{amsmath}
\usepackage{array}
\usepackage{wrapfig}
\usepackage{colortbl}
\usepackage{xcolor}
\usepackage{pifont}
\usepackage{makecell}
\usepackage{lscape}
\usepackage{longtable}
\usepackage{float}

\begin{document}

\pagestyle{fancy}
\rhead{\includegraphics[width=2.5cm]{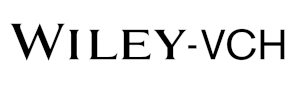}}

\title {Fairness in Federated Learning:  
Trends, Challenges, and Opportunities}
\centering
\maketitle

\author{Noorain Mukhtiar}
\author{Adnan Mahmood}
\author{Quan Z. Sheng}

\begin{affiliations}
Noorain Mukhtiar, Dr. Adnan Mahmood, Distinguished Prof. Quan Z. Sheng\\
School of Computing,  
Macquarie University, Sydney, NSW 2109, Australia\\
Email: noorain.mukhtiar@hdr.mq.edu.au, \{adnan.mahmood, michael.sheng\}@mq.edu.au
\end{affiliations}

\justifying
\begin{abstract}
At the intersection of the cutting-edge technologies and privacy concerns, Federated Learning (FL) with its distributed architecture, stands at the forefront in a bid to facilitate collaborative model training across multiple clients while preserving data privacy. However, the applicability of FL systems is hindered by fairness concerns arising from numerous sources of heterogeneity that can result in biases and 
undermine a system's effectiveness, with skewed predictions, reduced accuracy, and inefficient model convergence. This survey thus explores the diverse sources of bias, including but not limited to, data, client, and model biases, and thoroughly discusses the strengths and limitations inherited within the array of the state-of-the-art techniques utilized in the literature to mitigate such disparities in the FL training process. We delineate a comprehensive overview of the several notions, theoretical underpinnings, and technical aspects associated with fairness and their adoption in FL-based multidisciplinary environments. Furthermore, we examine salient evaluation metrics leveraged to measure fairness quantitatively. Finally, we envisage exciting open research directions that have the potential to drive future advancements in achieving fairer FL frameworks, in turn, offering a strong foundation for future research in this pivotal area.
\end{abstract}

\raggedright

\justifying
\section{Introduction}

\begin{figure} [!tb]
\centering
    \includegraphics[width=1\textwidth, trim = 0cm 4cm 0cm 0cm, 
    clip]{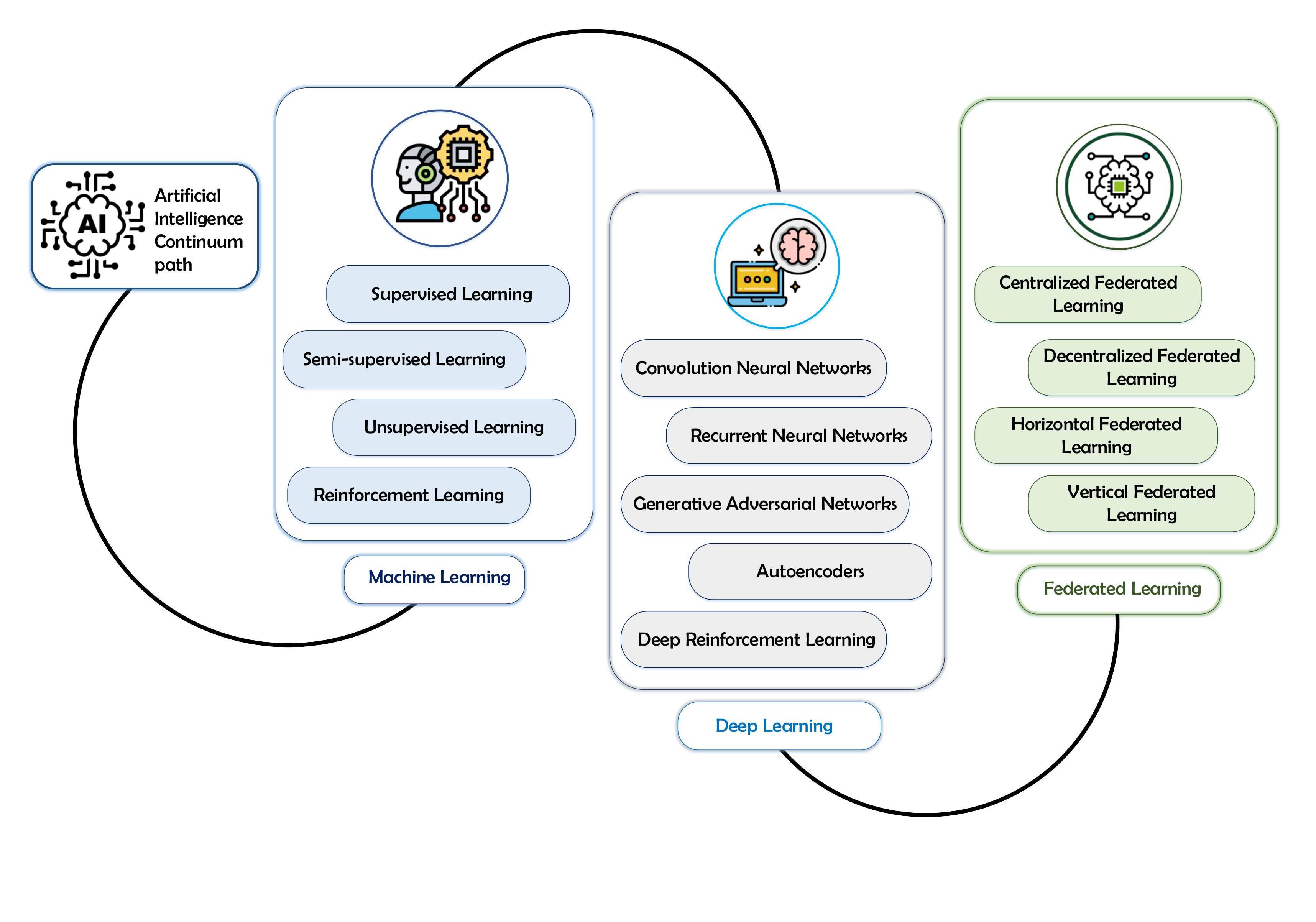}
    \caption{The evolution of artificial intelligence; from machine learning to federated learning.}
    \label{AI roadmap}
\end{figure}

The rapid evolution of Artificial Intelligence (AI) marks a new epoch in technological innovation hence revolutionizing the way machines perceive and interact with the world. The AI continuum evolves from basic algorithmic decision-making to sophisticated learning systems, i.e., as broadly categorized in \textbf{Figure \ref{AI roadmap}}.  {\em Machine Learning (ML)}, a branch of AI, automatically learns meaningful patterns and relationships from data without the need to be explicitly programmed. Another key branch, {\em Deep Learning (DL)}, is capable of processing vast amounts of data through multi-layer neural networks. Despite these remarkable advancements, ML and DL face several limitations, i.e., the need for large, centralized datasets and the risk of data privacy breaches.

Federated Learning (FL) addresses the above-mentioned challenges by enabling decentralized model training across multiple devices, a.k.a. clients or participants, and via preserving the data privacy while harnessing a collective intelligence of distributed data sources \cite{soltani2022survey, ijcai2024p919}. Unlike traditional centralized learning methods, FL allows each participating device to train a model locally using its own data, thereby ensuring that the raw data never leaves the device. Instead, only the updated parameters of the locally trained models are shared with a central server, wherein they are aggregated to constitute a global model. Consequently, this minimizes communication cost and transmission latency, and addresses several critical data issues, including but not limited to, data access, security, and privacy \cite{Vucinich}. 

Besides numerous advantages, FL faces several challenges that hinder its practical deployment and impact the efficiency and effectiveness of the learning process. Among other challenges, {\em fairness} has emerged as one of the most significant concerns. The inherent heterogeneity in FL, stemming from variations in client resources, data distributions, network conditions, and participation frequencies, creates disparities in model contributions leading to  {\em biased} learning outcomes. One of the key sources of unfairness in FL arises from client selection strategies, wherein clients with stronger capabilities are often favored while resource-constrained clients are either excluded or underrepresented. This selective participation results in biased global models that disproportionately favor certain user groups while neglecting others, exacerbating algorithmic disparities. Additionally, other factors, i.e., training instability, aggregation mechanisms, and optimization objectives, also influence fairness making it a multifaceted challenge requiring tailored solutions.

Given the significance of fairness in FL, numerous techniques have already been proposed to mitigate bias. These approaches inherently involve trade-offs among fairness, accuracy, efficiency, and privacy, thereby making it quite challenging to balance such competing objectives in practice. Moreover, distinct fairness notions adopted in fairness-aware FL algorithms introduce additional complexities by leveraging different optimization strategies, impacting model convergence, and affecting how biases are mitigated across the heterogeneous clients. Hence, the evolving nature of fairness-aware FL necessitates a systematic and comprehensive analysis of existing strategies, their effectiveness, and limitations. Several surveys have been published thus providing valuable insights into fairness in FL, covering its fundamental principles, and key challenges.  For instance, \cite{Vucinich} presents an overview of fairness challenges offering insights into the key issues of the field, \cite{Shi} provides a broad exploration of fairness-aware FL outlining general methodologies and concerns, and \cite{chen2023privacy} offers a unique perspective by examining fairness in context of its trade-off with privacy. Whilst these surveys provide a strong foundation of fairness in FL, an in-depth analysis of the root causes of bias, structured categorization of fairness-aware techniques, and comprehensive evaluation of their strengths and limitations are still required. We, therefore, address these gaps by exploring various sources of bias, analyzing state-of-the-art fairness techniques, and discussing multidisciplinary fairness applications across various domains. Additionally, we explore fairness evaluation metrics in detail and outline exciting future research directions, in turn, offering a more holistic perspective on fairness in FL.

\subsection{Federated Learning -- Fundamentals and Variants}

\subsubsection{Federated Learning Architectures}
\begin{figure} [!tb]
\centering
    \includegraphics[width=1\textwidth, trim = 0cm 2cm 0cm 2cm, 
    clip]{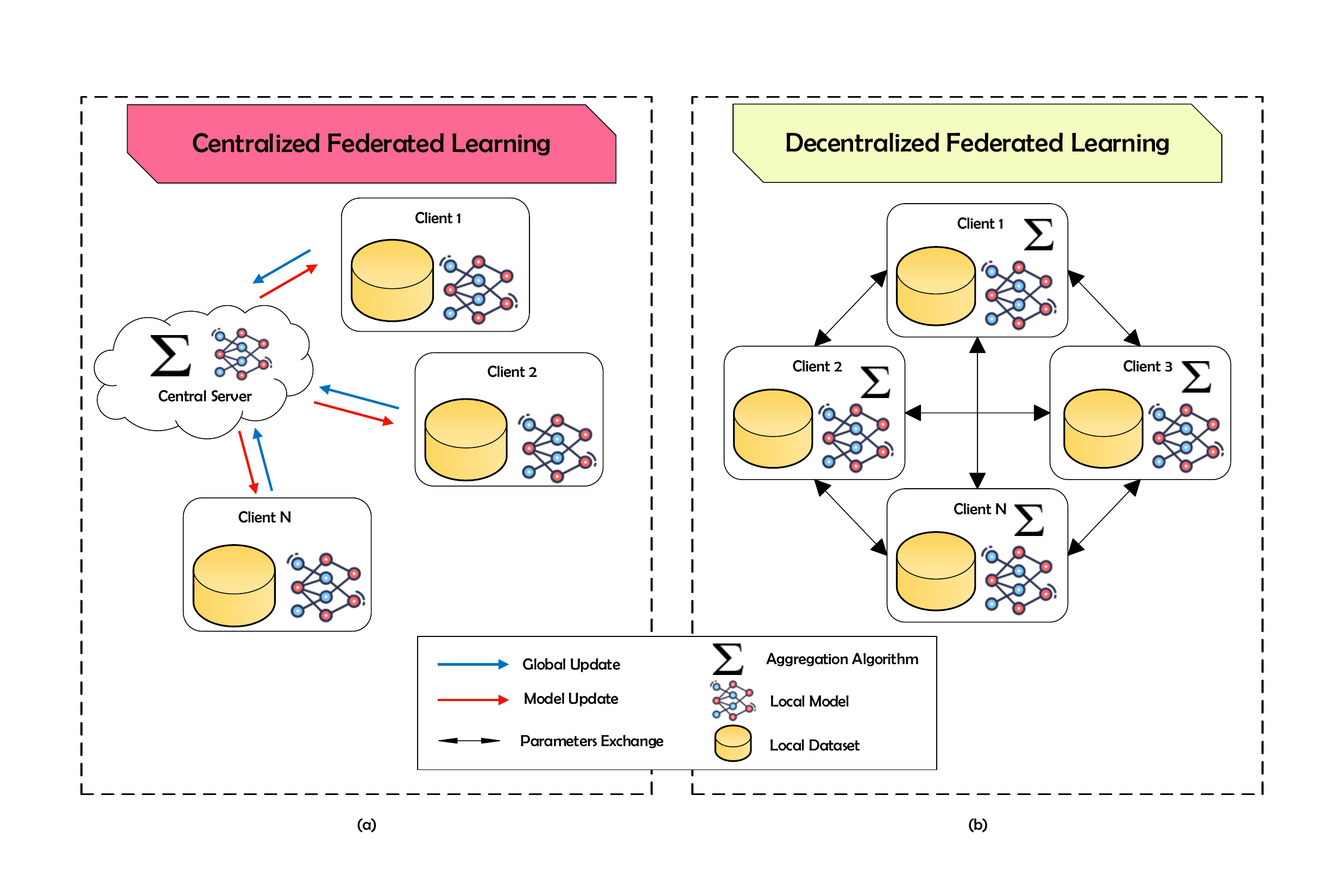}
    \caption{Centralized and decentralized federated learning architectures.}
    \label{CSD architecture}
\end{figure}

The two main FL architectures include centralized and decentralized federated learning, both differing in the way (a) they are structured, (b) handle client data, and (c) learn the model.
A \textit{Centralized Federated Learning (CFL)} based architecture implicates the collaboration of a central server and multiple FL participants to train a global model \cite{rafi2024fairness}. In this type of architecture, each client trains its own local dataset independently and sends updated parameters to the central server. Subsequently, the server aggregates the collected model parameters from each participating client, constitutes a consolidated central model, and redistributes to the clients, as depicted in \textbf{Figure \ref{CSD architecture} (a)}. 
This iterative process preserves privacy by keeping data on local devices while leveraging the collective updates of the participating clients. In contrast, a \textit{Decentralized Federated Learning (DFL)} based architecture is fully autonomous where clients can independently communicate with each other, thereby, eradicating the requirement of central server \cite{decentralized2023}, illustrated in Figure \ref{CSD architecture} (b). 
Such type of architecture allows clients to choose one or more peers for data transmission, enabling a bidirectional mode of communication, with either fixed or dynamic interconnections. Each client maintains and improves its model by incorporating knowledge from neighboring clients, increasing resilience to failures by eliminating dependence on a single point of control.

\subsubsection{Types of Federated Learning} 
Based on data partitioning schemes, FL framework can be segregated into three main types, i.e., horizontal FL, vertical FL, and federated transfer learning, with each delineating how data is distributed and utilized across multiple clients. These schemes significantly impact the design and structure of FL models, determining whether data is partitioned based on features, samples, or a combination of the both. 
\textit{Horizontal Federated Learning (HFL)} is applied when datasets across clients share the same feature space but different samples. In HFL settings, each client's dataset possesses significant overlap in features but minimal overlap in users \cite{huang2022fairness}. This approach leverages the common feature dimensions among participants, taking data with similar characteristics but from different users for joint training. Similarly, \textit{Vertical Federated Learning (VFL)} is adopted where datasets share the same sample space but different feature space. In VFL settings, the overlap in users across client datasets is higher than the overlap in data features, i.e., the same users appear in multiple datasets, however, their associated features vary \cite{wen2023survey}. \textit{Federated Transfer Learning (FTL)} is employed when datasets differ in both sample space and feature space. FTL leverages transfer learning techniques to map distinct feature spaces into a unified representation, consequently, using them for training on local datasets \cite{Asadullah}. 

\subsection{Fairness -- A Core Challenge in Federated Learning}
The existing FL approaches can be broadly categorized into {\em performance-oriented} and {\em fairness-oriented} approaches. 
\subsubsection{Performance-oriented Approaches} 
Performance-oriented approaches in FL prioritize the enhancement of (either or both of the) model performance \cite{zhao2022participant}, \cite{OORT}, \cite{FedProx}, \cite{AAAImodelperformance} and convergence speed \cite{OORT}, \cite{FedCS}, \cite{convergencespeed} without due consideration for the interests of individual clients. These approaches focus on optimizing the global model's performance by choosing the best-performing clients or, in some scenarios, randomly selecting clients \cite{mcmahan2017communication,FedDyn}.
Consequently, clients with lesser capabilities might be excluded from the FL training process. Such approaches often leverage threshold-based criteria (i.e., transmission speed, bandwidth availability, and local accuracy) in a bid to filter out less qualified clients and select those deemed to be of higher quality. However, this practice can inadvertently lead to oversampling of clients from specific groups,  where high-performing clients may be favored repeatedly during the FL training process while neglecting the contributions of others, resulting in unfair client selection and consequent deterioration in model performance. Moreover, these approaches risk depriving poor-quality clients, restricting them from participation in FL training, thereby, driving them to leave the system. Whilst performance-oriented approaches could achieve rapid and high-quality model convergence, they venture into creating imbalances undermining the inclusivity and fairness of FL models.

\subsubsection{Fairness-oriented Approaches} 
The continued participation of clients is essential to maintain the long-term sustainability of an FL system. Fairness-oriented approaches in FL ensure that all clients are treated equitably throughout the learning process. It is pertinent to mention that fair client selection does not imply picking everyone with equal probabilities, rather it encourages to provide equal opportunities for all clients to participate and contribute to the learning process. To achieve fairness in client selection, fairness-based approaches require striking a balance between the interests of both FL clients and the server. These methods emphasize the inclusion of underrepresented or disadvantaged clients by adjusting client selection strategies, model aggregation techniques, and resource allocation to mitigate biases and ensure that every participant contributes to, and benefits from, the global model. By incorporating mechanisms that balance the influence of all clients, fairness-oriented approaches aim to produce a more representative and unbiased model. This can involve strategies such as equalizing the weight of client updates, compensating for data imbalances, or ensuring that even clients with limited computational power or less frequent participation are not marginalized.

\subsection{Contributions}
This survey focuses on the key aspects and approaches leveraged to foster fairness in the dynamic FL environments. We provide a comprehensive overview of the state-of-the-art literature in a bid to address the identified gaps not extensively explored. The main contributions of our survey are delineated 
in answering the questions 
as follows:
\begin{itemize}
\item [Q1:] 
{\em What are the key notions of fairness in the context of FL and how can they be harmonized to ensure fair client selection across diverse participants?} 
\item [Q2:] {\em What are the fundamental sources of bias in FL and how do they affect the model performance?}
\item [Q3:] {\em What are the strengths and limitations inherent to the state-of-the-art fairness-aware strategies, implemented to mitigate bias across heterogeneous client populations?} 
\item [Q4:] {\em How fairness is integrated in FL-based multidisciplinary applications?} 
\item [Q5:] {\em What are the current research trends adopted to quantitatively evaluate fairness in the FL environment?} 
\item [Q6:] {\em What are the critical open research directions to foster the development of more inclusive and equitable FL frameworks?} 
\end{itemize}

\begin{figure} [!bt]
\centering
    \includegraphics[width=1.06\textwidth, trim = 4cm 10cm 0cm 11cm, 
    clip]{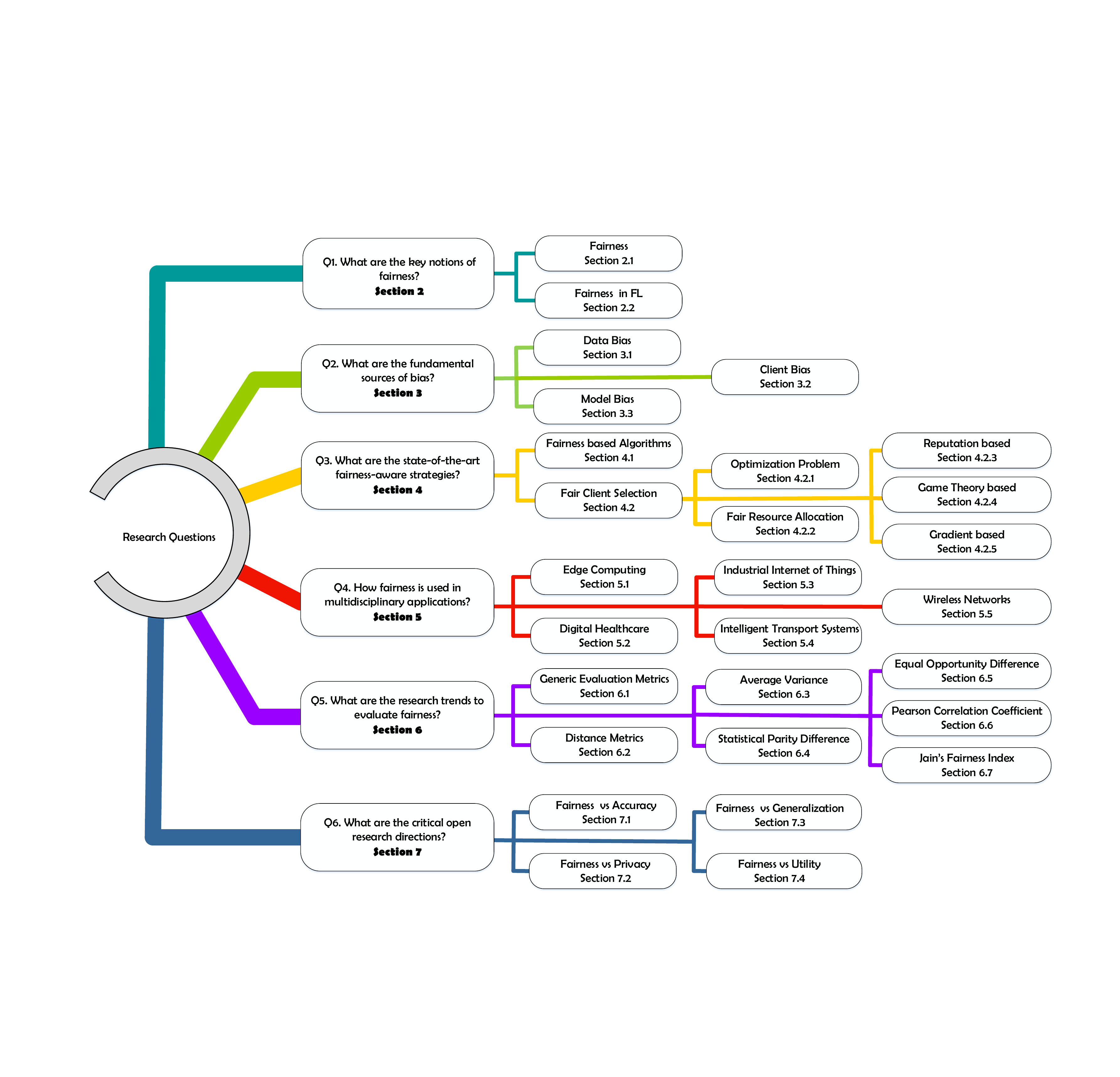}
    \caption{Taxonomy of this survey; research questions and corresponding sections addressing them.}
    \label{taxonomy}
\end{figure}

The taxonomy illustrating the above research questions and section-wise representation of their respective answers is given in \textbf{Figure \ref{taxonomy}}. 
Section \ref{Sec:fairness} 
interprets the fundamental perspectives and attributes of fairness. 
Section \ref{Sec: The roots of Bias} explores the potential roots of biases in FL and their inherent impacts on model performances. 
Section \ref{Sec: FAFL} categorically segregates the fairness-aware strategies based on the techniques involved, providing deep insights into their strengths and limitations. Section \ref{Sec: Cross-domain fairness applications} recognizes the challenges encountered and methods to mitigate those challenges while maintaining fairness across multidisciplinary FL-based applications. 
Section \ref{Sec: Fairness Evaluation Metrics} evaluates the adoption trends of fairness evaluation metrics in existing research. Section \ref{Sec: Open Research Directions} identifies potential open research directions within the realm of bias mitigation in FL systems. 
Finally, Section \ref{Sec: Conclusion} provides some concluding remarks.

\subsection{Paper Selection}
The articles referenced in this 
paper are high-quality publications sourced from reputable transactions (i.e., IEEE Transactions on Information Theory, IEEE Transactions on Neural Networks and Learning Systems, IEEE Transactions on Parallel and Distributed Systems), leading journals (i.e., ACM Computing Surveys, Computer Networks, Electronic Markets, and Internet of Things), and top-tier conferences, including but not limited to, ICML, ICCV, NeurIPS, WWW, KDD, SynSys, OSDI, ICLR, CVPR, IJCAI, ICDE, and AAAI. Initially, the article selection process utilized search strings, i.e., ``fair'' or ``fairness-aware'' + ``Federated Learning'' or ``Decentralized Learning'' across prominent resource libraries, including ACM, IEEE, Springer, Google Scholar, and ScienceDirect. Subsequently, the selected articles were categorized based on their publication in leading journals, top-tier conferences, and year of publication. Additionally, early access papers were included from mentioned sources as well as arXiv. The final selection was guided by the quality, novelty, and relevance of the proposed research to the scope of this study.

\section{Fairness}
\label{Sec:fairness}

Fairness is considered as a fundamental aspect in today's
fast-paced and technologically revolutionized world. As a concept, it has a range of interpretations in varying contexts and across diverse disciplines. From ethical principles to complex systems, it serves as an essential pursuit to ensure equitable treatment for everyone. In this section, we delineate the broad notions of fairness, without and within the specific context of FL, enlightening their implications and intersections.

\subsection{Fairness -- Perspectives and Attributes} \label{SubSec:Fairness}
In terms of social choice theory, utilitarianism and egalitarianism are two prominent opinions. Utilitarian fairness aims to optimize the overall utility of the society, whereas, egalitarian fairness focuses on ensuring maximum benefit to the badly off individuals \cite{Guojun}. Within the realm of decision-making and AI, fairness is predominantly segregated into seven categories \cite{Chen}. 

More specifically, \textit{individual fairness} \cite{Dwork,chu2021fedfair} implies that for a particular task, any two individuals possessing similarities should be categorized in a similar manner. \textit{Group fairness} \cite{Vucinich}, on the other hand, is associated with challenges encountered during the quantification of sensitive attributes (i.e., race, gender, and age) within groups and mitigation of inherent biases that might occur while dealing with them. \textit{Equality of opportunity} \cite{hardt2016equality} prioritizes fairness within the groups assuring that individuals with similar outcomes are treated equally by the predictive model irrespective of their differences in other attributes. \textit{Disparate impact} \cite{Feldman} reveals that AI systems, whether intentionally or unintentionally, have an undue adverse effect on certain groups. \textit{Fairness through unawareness} \cite{Giandomenico,Jiahao} states that an algorithm is considered to be fair as long as it does not explicitly utilize any protected attributes in the decision-making process. \textit{Disparate treatment} \cite{Wang} occurs when ML model generates varying decisions for individuals based on legally protected and sensitive attributes. Finally, \textit{subgroup fairness} \cite{Changjian} emphasizes the equitable treatment of particular groups or segments within larger demographic categories. 

It is pertinent to mention that these diverse dimensions of fairness are not mutually exclusive and may overlap in practical scenarios. The pursuit of fairness in AI and the decision-making process is a personalized solution and demands meticulous consideration of the intricacies of the factors involved \cite{LiuZ0HGS24}.

\subsection{Fairness in Federated Learning} 
\label{SubSec: Fairness in FL}
Federated Learning, characterized by its distributed structure and decentralized data sources, encounters unique challenges and opportunities regarding fairness. In this section, we will discuss the broad notions of fairness in terms of federated learning frameworks as adopted in existing literature. 

\textit{Client-level Fairness} \cite{Shi,Pengyuan} ensures equitable participation of all clients involved in the FL process. This notion aims to mitigate potential biases or disparities by prioritizing the inclusion of underrepresented or unrepresented participants. 
\textit{Group Fairness} \cite{Shi,ezzeldin,ijcai2024_EABFL} maintains equity among different demographic groups of clients participating in federated training. This concept focuses on alleviation of biases in the trained model's performance against specific groups based on sensitive attributes, i.e., age, gender, and race. 
\textit{Accuracy Parity or Performance Distribution Fairness} \cite{rafi2024fairness,Asadullah,Shi} measures the uniformity of performance distribution across FL clients. The main idea is to achieve a correct balance between fairness and accuracy while ensuring comparable performance levels for each client.
 
\textit{Good-Intent Fairness} \cite{li2020fair_ICLR,mohri2019agnostic} aims to reduce the maximum loss among protected groups, thereby preventing the overfitting of any single model at the cost of others. \textit{Contribution Fairness} \cite{Xinyi,Lingjuan} ensures the distribution of rewards in accordance with client's contribution (i.e., uploaded parameters or gradients). This notion prioritizes rewarding clients with the maximum contribution in FL training rather than those contributing negligibly. \textit{Regret Distribution Fairness} \cite{rafi2024fairness,Shi,yu} reduces the discrepancy among FL clients' regret on waiting times to obtain incentive payouts, considering the duration owner has waited to receive full payoff. \textit{Expectation Fairness} \cite{Shi,yu} is rooted with the principle of regret distribution fairness. This notion aims to establish fairness by minimizing the inequality among clients over time as incentive rewards are incrementally disbursed. 

Each of these mentioned fairness notions in FL brings forth unique advantages across diverse scenarios. Fairness-oriented strategies within an FL framework address systematic biases in data distribution and client participation by developing algorithms that foster fair representation and decision making thus ensuring uniform and unbiased model performance among diverse clients. 

\section{The Roots of Bias in Federated Learning} 
\label{Sec: The roots of Bias}
\begin{figure} [!bt]
\centering
    \includegraphics[width=1\textwidth, trim = 0cm 4cm 0cm 1cm, 
    clip]{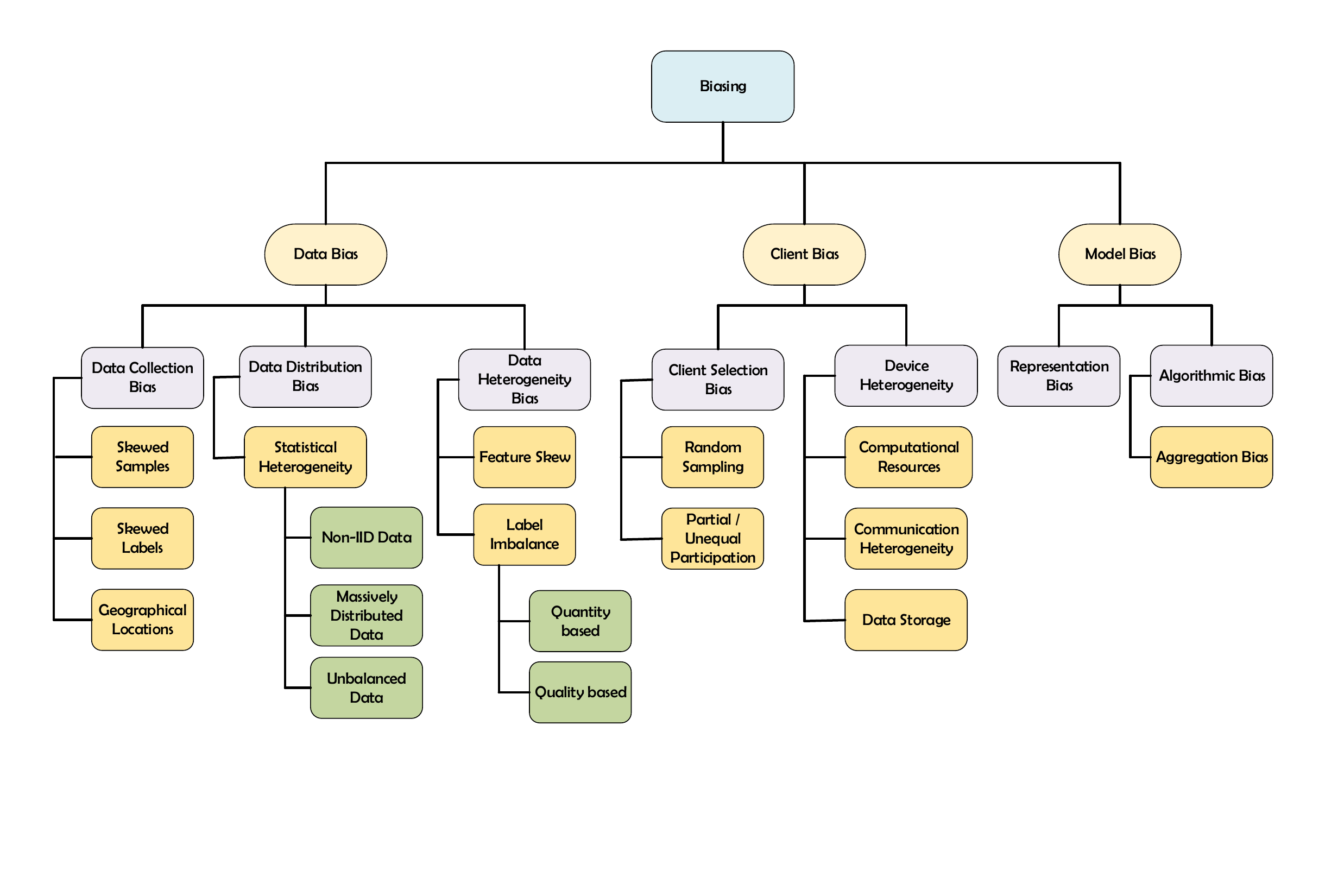}
    \caption{The roots of bias in federated learning.}
    \label{roots of bias}
\end{figure}

Federated Learning, with the assumption of homogeneous data distribution and network structure among participants, has demonstrated remarkable progress in the field. However, in practical scenarios, this framework is not immune to the considerable discrepancies posed by the substantial levels of FL training process. The roots of unfairness in FL models, as mentioned in the literature, span a variety of factors. Among such, the key causes include \textit{Data, Client, and Model biases}, illustrated in \textbf{Figure \ref{roots of bias}}. We will discuss the details in this section. 

\subsection{Data Bias} 
\label{SubSec: Data Bias}
Data bias can manifest at various stages, i.e., from data collection and distribution to the inherent heterogeneity within datasets. \textit{Data Collection Bias} \cite{Selialia} occurs due to heterogeneity between data collection devices' specifications and their respective geographical locations, thereby introducing inequity in model performance across clients using a shared global model. \textit{Skewed Labels Bias} \cite{ezzeldin} emerges from variation in data collection environment, representing diverse label distributions across participating clients.

\textit{Skewed Sampling} \cite{Venkataraman} takes place due to the uneven allocation of data samples among different groups in a dataset. FL model training relies on data distributions among edge devices (clients) to collaboratively train global models. Conventional aggregation algorithms i.e., weighted averaging \cite{mcmahan2017communication} might result in poorly averaged results and low model convergence speed due to \textit{Data Distribution Bias} \cite{Ou}. \textit{Statistical heterogeneity} \cite{ye2023heterogeneous} refers to the variation in data distribution across different clients. In FL settings, it can be further classified as \textit{Non-Independent and Identically Distributed (Non-IID), Massively Distributed, and Unbalanced Data}. In Non-IID data settings, each client's data does not reflect the overall data distribution due to its non-independent and identical nature. For massively distributed data, the ratio of FL clients is much larger than the average number of data points across each client. Unbalanced data bias results in a non-uniform distribution of data points across participants, due to diverse usage data patterns, producing different data sizes. The word ``unbalance'' refers to the variation in the number of samples per device or category.

\textit{Data Heterogeneity} encompasses differences in features, scales, formats, and labeling patterns within the datasets. \textit{Label Imbalance} \cite{Zhu} represents the diversity in label distribution varying from client to client, encompassing a distinct aspect of data variation. It can be further classified as quantity-based and distribution-based label imbalance. \textit{Quantity-based Label Imbalance} occurs when each participant owns data samples with a fixed number of labels. In \textit{Label Distribution Bias}, each participant is assigned a proportion for each label in accordance with Dirichlet distribution. \textit{Feature Skew} \cite{Venkataraman,Qinbin} implies that the feature distribution of local data among individual clients varies significantly. This phenomenon usually occurs in complex real-time environments.

\subsection{Client Bias} 
\label{SubSec: Client Bias}
Client bias can exert an additional layer of complexity to the fairness challenge. During FL training, only a few devices (clients) are chosen for participation in each round, resulting in \textit{partial/unequal client selection} \cite{Smestad}. Most of the existing client selection algorithms prioritize the selection of the fastest client, which might boost the training process. However, this approach unfairly excludes the clients with less priority, thereby depriving their chances to participate simultaneously. This \textit{biased client selection} \cite{huang2020efficiency} might overlook certain data segments and involve fewer data points, therefore, compromising data diversity and performance of the model training. Moreover, \textit{Random Client Selection} exhibits the risk of bias by potentially picking clients with over-represented data.

\textit{Device Heterogeneity Bias} \cite{ye2023heterogeneous} emerges due to uneven contribution from devices with diverse levels of \textit{computational resources}, \textit{network connectivity}, \textit{data storage}, and \textit{battery life}. Clients with higher capabilities contribute more in FL training process rather than those with limited resources, potentially causing faults and inactivation of some participating nodes (i.e., stragglers). This, in turn, results in a model that fails to fully represent the entire population, favoring clients with better resources. In distributed network settings, devices are typically deployed in varying network environments and connectivity settings (i.e., 3G, 4G, 5G, WiFi), introducing inconsistent communication bandwidth, latency, and reliability (a.k.a. \textit{communication heterogeneity}). 
During FL training, some devices might experience slow connection or even get disconnected due to network bandwidth and energy constraints, such disparities skew the communication dynamics and impact the effectiveness of the model. 

\subsection{Model Bias} 
\label{SubSec: Model Bias}
Model bias can arise due to statistical heterogeneity among participants in FL training, thereby resulting in biased datasets that do not encompass all labels. When the global model is trained on these locally biased datasets, it tends to form biased representations \cite{Zhang}. This phenomenon entails the emergence of client-specific clusters within the model's learned representations. \textit{Algorithmic Bias} \cite{chen2023privacy} refers to the disparities in the algorithm decision across distinct groups of clients defined by sensitive attributes. 
The fusion algorithms employed by the aggregator to combine the model updates from various clients may induce
\textit{aggregation bias} due to averaging techniques, i.e., equal or weighted averaging. 
FL algorithms become unfair by giving more weight to the contributions from large data populations, potentially amplifying effects of over or under-representing specific groups in a dataset \cite{Annie}.
\section{Fairness-Aware Algorithms -- An Overview of the State-of-the-Art Approaches in Federated Learning} 
\label{Sec: FAFL}
Biasing, if not addressed effectively, can exert a profound impact on both the server(s) and the client(s) side in an FL environment. It is pertinent to mention that unfair treatment can provoke distrust and dissatisfaction in clients, thereby discouraging their participation in the FL training process. Furthermore, treating all clients in an equal manner regardless of their respective contributions, can reduce the servers' ability to attract high-quality clients. This, therefore compromises the effectiveness of the FL models and results in models which may not generalize well. 

\subsection{Fairness-based Algorithms} 
\label{SubSec: Fairness-based algorithms}
Over the years, significant contributions have been made to tackle fairness-related challenges in the FL paradigm. Existing algorithms can be broadly classified into three major categories, i.e, \textit{Pre-Processing}, \textit{In-Processing}, and \textit{Post-Processing}, each aiming to address specific challenges related to fairness in FL settings.   

Pre-processing algorithms are employed prior to the model training phase in a bid to mitigate bias and enhance fairness in the training data. Such algorithms \cite{rafi2024fairness,Annie,Calmon} aim to rectify biases at the data level by leveraging intelligent client selection encompassing (via either or both of the) resource allocation and data heterogeneity management, to ensure equitable representation and participation amongst the clients. Accordingly, data is pre-processed to generate less biased dataset by either (a) modifying raw data samples, (b) altering the value of sensitive attributes, or (c) assigning weights to data samples. 

In-processing algorithms are utilized to customize learning algorithms or objective functions to integrate fairness in the FL model. These algorithms \cite{Chen,SUN} modify fairness at an algorithmic level by the addition of either (a) bias mitigation constraints, (b) discrimination aware regularizers, or (c) adversarial debiasing to the optimization formulation. Among such, the \textit{fairness constraints} method integrates fairness constraints to the loss function during FL training. The \textit{discrimination aware regularization} method adds a regularizer to the loss function; whereas the
\textit{adversarial debiasing} method simultaneously develops a predictor and an adversary in a bid to reduce the ability to predict the protected attribute from the predictive output.

Post-processing algorithms are applied to adjust the output of trained FL models and reduce bias after the aggregation of global model. This adjustment to the model prediction is made under specified fairness constraints to ensure that model decisions are fair and unbiased. Such algorithms \cite{rafi2024fairness} do not require changes to the model training process and operate without access to sensitive local data, thereby modifying outcomes instead of altering the classifier or training data. Some most common post-processing bias mitigation algorithms include (a) Equalized Odds (EO), (b) Calibrated Equalized Odds (CEO), and (c) Rejection Option Classification(ROC).

\subsection{Fairness-Aware Strategies in Federated Learning} 
\label{SubSec: Fairness-Aware Strategies in FL}
\begin{figure} [!bt]
\centering
    \includegraphics[width=0.85\textwidth]{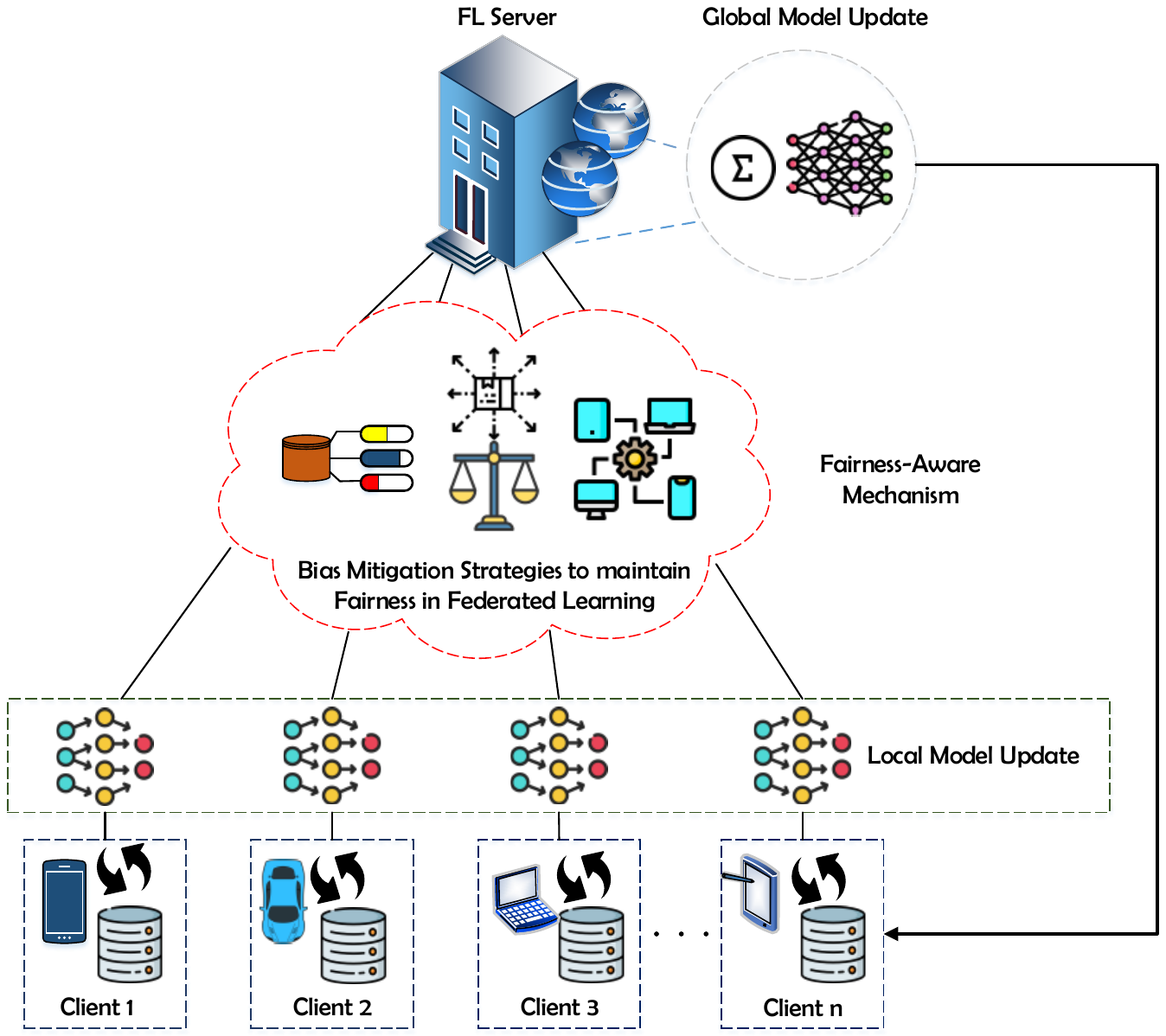}
    \caption{Fairness-aware strategies in federated learning.}
    \label{Fair Stretegies}
\end{figure}

One of the key sources of unfair treatment in an FL model lies in the client selection strategies. During federated model training, the task requester(s) submit different types of tasks (with or without specific task requirements) to the FL service provider (i.e., server). As illustrated in \textbf{Figure \ref{Fair Stretegies}}, the service provider then picks a fraction of clients, satisfying task requirements, from the available client pool to participate in FL training.  The selected participants involve their local data as well as communication and computational resources. Due to various heterogeneity factors among clients, their contribution, i.e., model performance and convergence speed, varies significantly. Fair client selection involves two stages, namely
(a) choosing \textit{initial client pool} for each FL task and (b) selecting \textit{subset of clients} for each FL round. It is pertinent to mention that even if a client is chosen to participate in FL training, it does not guarantee its participation in every FL round due to several factors, i.e., communication and computation cost, conflicting schedule, low battery, or unstable network connectivity.

Algorithmic bias in FL extends beyond client selection, influencing various stages of the learning process, from data pre-processing to model aggregation and decision-making. Bias can emerge due to imbalanced data distributions, heterogeneous client capabilities, or disparities in model updates, which may result in unfair outcomes for certain clients or groups. These biases can amplify disparities in performance, utility, and resource allocation, thereby affecting the overall fairness of FL systems. Addressing fairness requires a holistic approach that not only ensures fairness in client selection but also incorporates fairness-aware strategies at multiple levels of the learning process.
In this section, we will review state-of-the-art fairness-oriented FL strategies. Each of them is designed to strike a balance between fairness and optimized model performance parameters, i.e., utility, accuracy, efficiency, and convergence speed. However, every approach may not excel across all mentioned parameters simultaneously, depicted in \textbf{Table \ref{Table1}}. Understanding the trade-off dynamics inherent to each algorithm is essential for selecting an appropriate fairness-aware scheme in FL.  

\subsubsection{Optimization Problem Formulation} 
\label{Optimization Problem Formulation}
An optimization problem formulation strategy is adopted to tackle bias-related challenges encountered during FL model training. It involves the formulation of a local/global objective function as an optimization problem while ensuring the target fairness constraints. A comprehensive multi-criteria-based fair client selection and scheduling scheme was introduced in \cite{Meiying}. The algorithm aimed to maximize model performance while maintaining fairness at both initial and per-round client selection stages. In the first phase, a client selection metric was defined based on certain factors, i.e., data quality, client resources, and behaviors. Subsequently, the client pool selection challenge was formulated as an optimization problem and addressed using a greedy selection algorithm. 

In \cite{shi2023fairness}, the research focused on maintaining a balance between model performance and fairness considerations in FL participant selection. The proposed client selection algorithm was based on the Lyapunov optimization problem, designed to dynamically adjust FL client selection probabilities aligned with their respective participation time, reputation, and contribution to the model performance. The reputation threshold filtering was neglected to promote fair treatment and clients were allowed to restore their reputations despite achieving poor performance. Similarly, the multi-objective optimization function was formulated to achieve Pareto-optimal trade-offs between balanced fairness and accuracy in an FL environment \cite{AAAI2024fairtrade}. The framework was designed to maintain both statistical and causal fairness notions to ensure its adaptability across various FL contexts. 

These approaches enable precise control over fairness constraints while improving accuracy and thus can be utilized for various fairness notions (as delineated in Section \ref{Sec:fairness}), thereby providing a flexible and interpretable framework for addressing bias. It is pertinent to mention that defining appropriate fairness constraints often requires domain-specific insights, however, overly strict formulations can degrade model performance and convergence speed. Furthermore, solving fairness-aware optimization problems can be computationally demanding, particularly, in heterogeneous FL environments with resource-constrained clients.

\subsubsection{Fair Resource Allocation} 
\label{SubSec: Fair Resource Allocation}
A fair resource allocation strategy is employed to ensure equitable resource distribution across FL participants encompassing heterogeneous specifications, i.e., computational capacity, data quality, and network limitations.
In \cite{albaseer2023fair}, the research presented a dual-phased participant selection and scheduling scheme for clustered multi-tasks FL, focusing on the improvement of model convergence speed while considering all data distribution. In the initial phase, an algorithm was developed to maintain correct clustering and fairness across participants by leveraging bandwidth reuse for slower clients, i.e., those who took longer time to train their models, and utilizing device heterogeneity to schedule participants based on their delays. In the second phase, the server clustered participants as per the pre-determined threshold values and stopping conditions. When a particular cluster of clients reached stopping criteria, the server picked clients with better resources and lower delay by applying a greedy selection algorithm.

The $q$-Fair FL \cite{li2020fair_ICLR} algorithm, motivated by fair resource distribution in wireless networks, proposed an optimization objective to encourage fair and uniform accuracy distribution among FL devices. The algorithm was designed to minimize a combined reweighted loss characterized by parameter $q$, thereby allocating greater weights to the devices with greater losses and vice versa. 
To address this, a communication-efficient technique \textit{q-FedAvg} was introduced, specifically tailored for federated networks. The research in \cite{li2021efficient} introduced a logarithmic fairness algorithm to solve the multi-model challenge in an FL environment. Such algorithm was developed to (a) leverage the clients' heterogeneous resources for parallel multi-model training, (b) enhance the overall training efficiency, and (c) maintain fairness across each model. To accomplish this, the authors propounded a logarithmic function to represent the interdependence between the model training accuracy and the number of participating clients based on measurement outcomes. Subsequently, multi-model training was formulated as an optimization problem to identify an assignment matrix that maximized the overall training efficiency and ensured logarithmic fairness among each model. 

These strategies maintain fairness by dynamically adjusting resource allocations based on client capabilities, reducing overall training time, and mitigating stragglers' impact. Nevertheless, fair resource allocation may lead to inefficiencies, as prioritizing weaker clients can slow down training convergence. Moreover, heterogeneous client capabilities create bottlenecks, e.g., disparities in data quality can degrade model performance despite equal resource distribution.

\subsubsection{Reputation and Regret-based Client Selection} 
\label{Reputation and Regret based Client Selection}
A reputation and regret-based client selection strategies are utilized to achieve fair client selection by leveraging participants' historical data, i.e., past contribution and performance metrics. Such techniques usually include a penalty factor to discourage poorly performing individuals, promote fairness, and enhance the training process. In \cite{huang2020efficiency}, the authors introduced a reputation-based participant selection algorithm based on Contextual Combinatorial Multi-Armed Bandit \begin{math} (C^2MAB)\end{math} method to estimate each client's model exchange time as per their historical reputation. The article concentrated on minimizing the average model exchange time between the server and each client while adhering to relatively flexible long-term fairness and system constraints. The fairness challenge was formulated as a Lyapnov optimization problem to improve FL clients' participation rates. A penalty factor was specified to balance the trade-off between the objective function minimization and fairness constraint satisfaction.  

The work in \cite{ACM2024privacy} introduced an optimization algorithm that enhanced privacy and fairness in FL environment by eradicating the need to share centralized global model and reduce the dependency among participants. Initially, the gradient descent optimization started with a set of discrete points that converged to another set near the global minimum of the objective function. Subsequently, each participant independently initiated its own private global model (termed the confined model) and collaboratively optimized it towards the optimal solution. The algorithm was based on a regret function, which quantified the difference between the loss function of confined gradient descent and the centralized model.  Similarly, \cite{hu2024fair} introduced a fairness-driven FL framework to ensure equitable model performance across different protected groups. The proposed method employed bounded group loss as a fairness criterion, setting an upper limit on loss for each protected group rather than enforcing strict equality of losses. The optimization framework was modeled as a saddle-point problem, ensuring both theoretical convergence guarantees and fairness constraints. It incorporated a regret-based mechanism with a no-regret bound analysis which guided the optimization process to minimize fairness constraint violations while balancing overall utility. 

These methods promote fairness by incentivizing consistent participation while discouraging malicious behavior. However, many struggle to adapt to new or intermittently participating clients, which can lead to their exclusion or delayed opportunities. Moreover, balancing fairness with model performance remains a concern, as strict regret-based mechanisms might overlook short-term performance gains \cite{YangZHWS23}.

\subsubsection{Fairness via Game Theory} 
\label{SubSec:fairness via Game Theory}

A game theory-based strategy is employed to ensure the fair treatment and appropriate reward distribution among participants (agents) in an FL system by leveraging basic concepts of game theory, i.e., Cooperative games, Nash equilibrium, and SV. The research in \cite{rankcorefed2024fair} investigated a novel fairness notion termed Proportional Veto Core (PVC) to ensure fairness in the utility distribution among participating agents. The algorithm aimed to achieve fairness by ensuring that the final model is PVC-stable, considering ordinal preferences of agents. It guaranteed fairness based on the models' ordinal rank rather than solely relying on their utility values.

In \cite{Fan_2022}, the authors debuted a metric based on SV to enhance fairness in FL settings. Such metric was designed by completing a matrix that comprised all possible contributions from various subsets of clients. The proposed measure was based on a solution of a low-rank matrix completion problem for the utility matrix. To address the practical scenarios, i.e., too large utility matrix, the Monte-Carlo sampling strategy was adopted to reduce both space and time complexities.
In \cite{shi2022fedfaim}, a fair incentive mechanism-based algorithm was introduced to ensure two notions of fairness, i.e., aggregation fairness and reward fairness. Aggregation fairness was achieved via effective gradient aggregation method. This method evaluated the quality of the local gradients based on the marginal loss value to filter out low-quality ones and subsequently aggregated local gradients as per their model updates. Similarly, reward fairness was attained through a contribution evaluation method based on the Shapley Value (SV), accompanied by a reward allocation technique. This technique relied on the reputation and distribution of local and global gradients, incorporating reputation scores to ascertain the performance of the model assigned to each participant. 

These approaches foster collaboration among heterogeneous clients, mitigate free-riding behavior, and provide mathematically grounded fairness guarantees. However, involves high computational complexity which can hinder real-time adaptability in large-scale FL settings. It is worth mentioning that some game-theoretic solutions often rely on theoretical constraints, assumptions, and extensive parameter tuning, thus limiting their applicability in practical deployments.

\subsubsection{Gradient-based Client Selection} 
\label{SubSec: Gradient-based Client Selection}
A gradient-based client selection strategy is applied to address fairness issues by adjusting the gradient updates during FL training process. Such strategy aims to design fair and accurate models by either incorporating fairness-aware regularization terms into the loss function or employing reweighting techniques to consider underrepresented population of clients. In \cite{chen2024fairCvpr}, the authors addressed performance fairness in FL under domain skew by resolving parameter update conflicts and model aggregation bias. Initially, the algorithm discarded insignificant parameter updates based on discovered characteristics to avoid poorly performing clients from being overwhelmed by those updates. Subsequently, a fair aggregation objective was incorporated to prevent global model bias towards certain domains, ensuring the continuous alignment of global model with an unbiased model. 

The work in \cite{Gradient_2025} introduced a multi-objective optimization algorithm designed to fine-tune global gradients in FL. It aimed at minimizing the average loss across all clients while reducing the gradient conflicts between the global and local gradients. Particularly, the global gradient was refined by incorporating a sub-optimization objective alleviating conflicts between the global gradient and the local gradient based on the largest deviation. Another work \cite{Reweight_2025} envisaged a local training algorithm that decomposed clients' training objectives into fair sub-objectives based on local data distributions. Moreover, it incorporated a self-aware aggregation mechanism to counteract bias propagation during model aggregation by employing a distance-based reweighting strategy to dynamically adjust aggregation weights for each client.

These methods provide a mathematically principled way to enforce fairness constraints without requiring explicit modifications to the FL process. Moreover, these techniques maintain compatibility with different FL aggregation schemes, e.g., FedProx \cite{FedProx} and FedAvg \cite{mcmahan2017communication}, making them highly versatile across various FL settings. Nonetheless, regularization-based methods often lead to non-convex optimization, thereby, increasing the risk of convergence instability, especially in highly heterogeneous environments. Similarly, reweighting strategies can amplify variance in gradient updates, which may hinder the model convergence and increase the risk of overfitting to minority clients. Furthermore, such mechanisms require careful tuning of hyperparameters, e.g., fairness coefficients and reweighting factors, which may demand extensive empirical validation.

\newcolumntype{M}[1]{>{\centering\arraybackslash}m{#1}}
\newcolumntype{P}[1]{>{\raggedright\arraybackslash}m{#1}}
\definecolor{green}{rgb}{0.0, 0.5, 0.0}
\newcommand{\cmark}{\textcolor{green}{\ding{52}}}
\setlength{\tabcolsep}{2pt}
\renewcommand{\arraystretch}{1.5}
{
\scriptsize
\begin{longtable} {M{18mm} M{7mm} M{17mm} M{37mm} M{4mm} M{4mm} M{4mm} M{5mm} M{7mm} M{4mm} M{15mm} M{15mm} M{15mm}} 
 \caption{Fairness-aware algorithms in federated learning.} \label{Table1}\\
    \hline
    \textbf{Technique Name and Reference} & \textbf{Year} & \textbf{Associated Algorithm(s)} & \textbf{Description} & \textbf{F} & \textbf{A} & \textbf{U} & \textbf{MP} & \textbf{MCT} & \textbf{E} & \textbf{Problem(s) Addressed} & \textbf{Outcomes} & \textbf{Dataset} \\ 
    \hline
    \endfirsthead
    \hline
    \textbf{Technique Name and Reference} & \textbf{Year} & \textbf{Associated Algorithm(s)} & \textbf{Description} & \textbf{F} & \textbf{A} & \textbf{U} & \textbf{MP} & \textbf{MCT} & \textbf{E} & \textbf{Problem(s) Addressed} & \textbf{Outcomes} & \textbf{Dataset} \\ 
    \hline
    \endhead
    \hline
    \multicolumn{13}{c}{\scriptsize Legends: F: Fairness, A: Accuracy, U: Utility, MP: Model Performance, MCT: Model Convergence Time, E: Efficiency} \\
    \hline
    \endfoot
    \raggedright FedMC \cite{Gradient_2025} & 2025 & \raggedright Fine-tuning global gradients & \raggedright Reduced the gradient conflict between the global and the local gradients & \cmark & \cmark && \cmark & \cmark && \raggedright Performance distribution fairness & \raggedright Fairness and convergence rate & CIFAR-10, CIFAR-100, and EMNIST \\ 
    \hline 
    \raggedright SFFL \cite{Reweight_2025} & 2025 & \raggedright Self-aware aggregation & \raggedright Designed self-aware aggregation method to reduce bias propagation during aggregation & \cmark & \cmark & & \cmark & & & \raggedright Inconsistent update objectives & \raggedright Fairness and privacy  & ACS (Employment, Income, and HealthInsurance) \\
    \hline
    \raggedright FairTrade \cite{AAAI2024fairtrade} & 2024 & \raggedright Multi-objective optimization & \raggedright Formulated the multi-objective optimization function to achieve Pareto-optimal trade-offs between balanced fairness and accuracy in FL environment & \cmark & \cmark &&&&& \raggedright Statistical and causal fairness & Fairness and accuracy & Adult, Bank, Default, Law, and KDD \\
    \hline 
     \raggedright CGD \cite{ACM2024privacy} & 2024 & \raggedright Objective function optimization &\raggedright Enhanced the privacy preservation in FL environment by eliminating the centralized sharing of global model while ensuring fairness & \cmark & \cmark &&& \cmark && \raggedright Priavte and equitable clients' participation & \raggedright Fairness, differential privacy, and model convergence rate & MNIST, CIFAR-10, Location 30, and Energy HEC \\
    \hline
    \raggedright Rank-Core-Fed \cite{rankcorefed2024fair} & 2024 & \raggedright Cooperative game theory & \raggedright Measured the quality of output models based on their ordinal rank rather than their cardinal utility  & \cmark && \cmark &&&& \raggedright Proportional veto core fairness & \raggedright Fairness and high utilitarian social welfare  & MNIST and CIFAR-10 \\
    \hline
    \raggedright FedHEAL \cite{chen2024fairCvpr} & 2024 & \raggedright Parameter update consistency & \raggedright Discarded insignificant parameter updates to neglect updates from poor performing clients & \cmark &&& \cmark && & \raggedright Parameter update conflicts and model aggregation bias & \raggedright Fairness and model generalization & Digits and Office-Caltech \\
    \hline
     \raggedright PPFL \cite{hu2024fair} & 2024 & \raggedright Saddle point optimization problem & \raggedright Employed bounded group loss as a fairness criterion, setting an upper limit on loss for each protected group & \cmark && \cmark & && & \raggedright Group unfairness & \raggedright Improved fairness and prediction accuracy & Communities and Crime, ACS PUMS, and CelebA \\
    \hline
    \raggedright Multi-criteria based client selection and scheduling \cite{Meiying} & 2024 & \raggedright Greedy selection algorithm & \raggedright Ensured fairness with multi-criteria based client selection and scheduling scheme & \cmark &&& \cmark &&& \raggedright Fair client selection & \raggedright Fairness and improved learning performance & MNIST and CIFAR \\ 
    \hline
    \raggedright FairFedCS \cite{shi2023fairness} & 2023 & \raggedright Lyapnov optimization & \raggedright Adjusted FL client selection probabilities aligning with their reputation, participation frequency, and contribution to the model performance & \cmark & \cmark &&&&& \raggedright Client selection bias & \raggedright Fairness and test accuracy & Fashion-MNIST and CIFAR-10 \\
    \hline
    \raggedright FEEL \cite{albaseer2023fair} & 2023 & \raggedright Clustered federated learning & \raggedright Ensured fairness and correct clustering among clients by leveraging bandwidth reuse of slower clients & \cmark &&&& \cmark & & \raggedright Fair client clustering & \raggedright Fairness and reduced training latency & FEMNIST and CIFAR-10 \\
    \hline
    \raggedright FedFAIM \cite{shi2022fedfaim} & 2022 & \raggedright Gradient aggregation and reward allocation techniques & \raggedright Maintained aggregation fairness and reward fairness using fair incentive mechanism technique & \cmark &&& \cmark &&& \raggedright Aggregation fairness and reward fairness & \raggedright Fair incentive mechanism & MNIST and CIFAR-10 \\
    \hline
    \raggedright COMFedSV \cite{Fan_2022} & 2022 & \raggedright Shapley values and contribution evaluation & \raggedright Improved the fairness of SVs which involved completing a matrix based on all possible contributions from various subsets of clients  & \cmark & & \cmark &&&& \raggedright Fairness for data evaluation in horizontal federated learning & \raggedright Fairness of federated SV & Synthetic, MNIST, Fashion-MNIST, and CIFAR-10 \\
    \hline
    \raggedright LFMB \cite{li2021efficient} & 2021 & \raggedright Optimization problem & \raggedright Formulated the multi-model training as an optimization problem to identify an assignment matrix to assure logarithmic fairness among individual models & \cmark &&&&& \cmark & \raggedright Client selection for parallel multi-modal training & \raggedright Fairness and efficiency & MNIST, Fashion-MNIST, and CIFAR-10 \\
    \hline
    \raggedright RBCS-F \cite{huang2020efficiency} & 2021 & \raggedright Lyapunov optimization and C$^2$MAB & \raggedright Introduced a reputation based client selection algorithm based on C$^2$MAB method to estimate each client's model exchange time as per their historical reputation while ensuring long-term fairness & \cmark & \cmark &&&& \cmark  & \raggedright Efficiency boosting client selection & \raggedright Improved training efficiency and long-term fairness & Fashion-MNIST and CIFAR-10 \\
    \hline
    \raggedright q-FFL \cite{li2020fair_ICLR} & 2020 & \raggedright Optimization function & \raggedright Designed to minimize a combined reweighted loss, characterized by parameter q which assigned higher weights to the devices with higher losses and vice versa & \cmark & & & & & \cmark & \raggedright Fair distribution of model performance & \raggedright Fairness, flexibility, and efficiency & Synthetic, Vehicle, Sent140, and Shakespeare \\ 
\end{longtable}
}

\section{Cross-domain Fairness Applications} 
\label{Sec: Cross-domain fairness applications}
\begin{figure} [H]
\centering
\includegraphics[width=1\textwidth, trim = 4cm 4cm 0cm 5cm, clip]{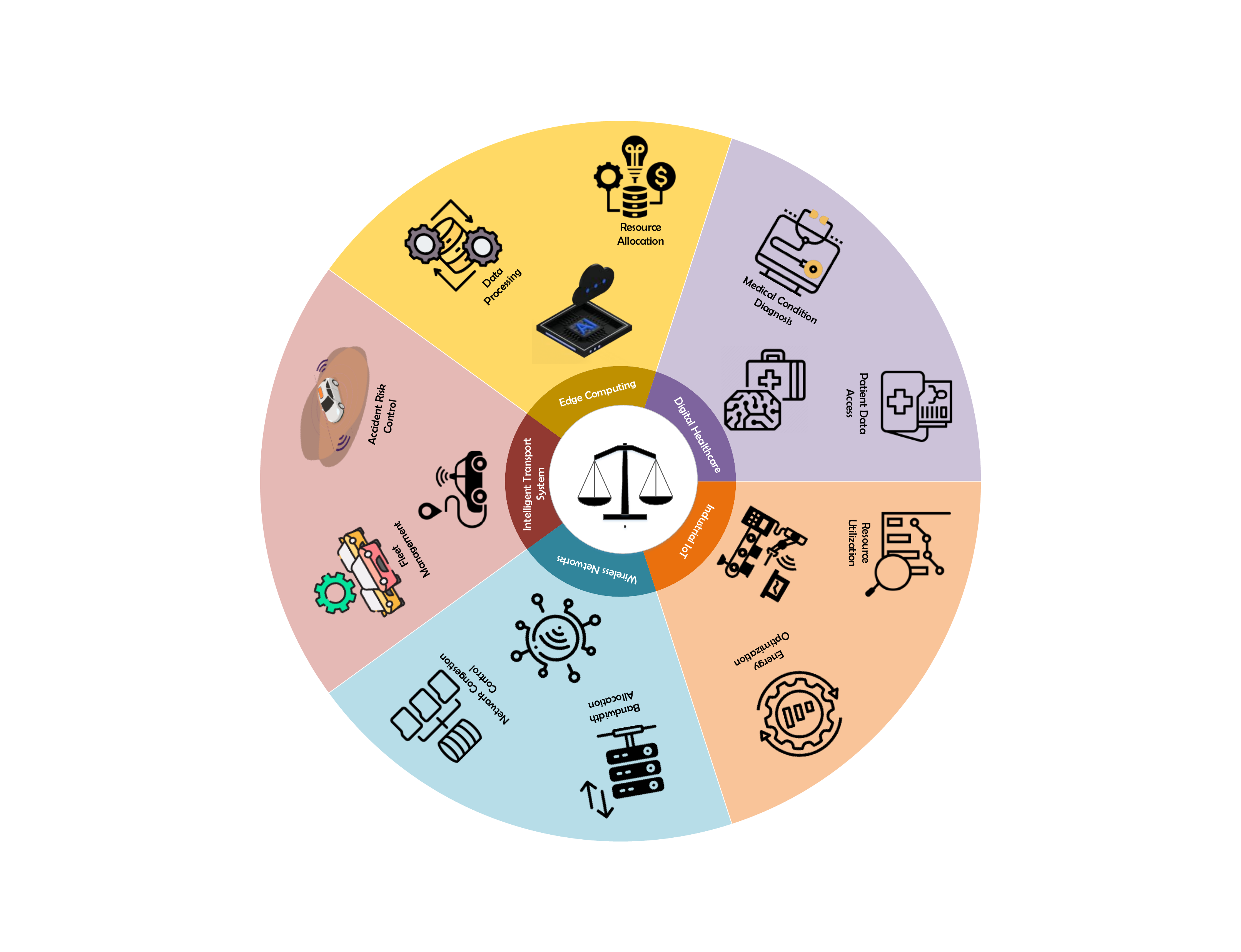}
    \caption{Multidisciplinary fairness applications in federated learning.}
    \label{Cross domain}
\end{figure}

Several researchers have investigated the applications of fairness-based FL environments across diverse domains, including but not limited to, edge computing (EC), digital healthcare, industrial internet of things (IIoT), wireless networks, and intelligent transport systems (ITS), thereby, highlighting its importance and implications. 
Such potential areas, as depicted in \textbf{Figure \ref{Cross domain}}, are discussed in this section. 

\subsection{Fair Resource Distribution in Edge Computing}
\label{SubSec: Fairness in Edge Computing}
In the dynamic realm of interconnected devices and real-time data processing demands, the emergence of EC has reshaped the way now data is being stored, processed, and managed. At its core, EC refers to a type of distributed computing where data storage and processing occur on connected devices located close to the edge of the network, where the proximity is determined by the system’s requirements \cite{edgecomp2022}. The idea revolves around reducing latency and bandwidth consumption by processing data near its source of generation. However, traditional ML approaches utilized in EC environments generally rely on centralized data aggregation for training, thereby, introducing significant privacy risks.

FL addresses the aforementioned challenge by (a) enabling decentralized model training across multiple edge devices without centralizing the data, (b) enhancing privacy by keeping data localized on edge devices, and (c) reducing the risk of sensitive information exposure \cite{zhang2021edge}. Since EC exhibits high system heterogeneity, the random participant selection method discards stragglers with lesser computing capabilities, consequently, leading to biased performance across participating devices. The research in \cite{mao2024federatededge} addressed the issue of unfair FL participant selection for EC applications. The authors envisaged a dynamic client selection algorithm that clustered clients into groups as per their respective computational efficiency. A data evaluation scheme was designed to estimate the importance of client datasets based on their local data distribution. Moreover, the DYNAMIC-SELECT algorithm was formulated to update local computational efficiency and data distribution parameters to regroup participants following periodic average aggregation. Their experimental evaluation illustrated that such algorithm improved client participation while enhancing local test accuracy and mitigating the performance bias of the global model across FL clients.    

\subsection{Bias Reduction in Digital Healthcare}
\label{SubSec: Bias Reduction in Digital Healthcare}
In this modern era of high-tech advancements, digital healthcare stands out as a reshaping force in the medical sector. It serves as an integration of digital technologies with healthcare to enhance the quality of patient care. This transformation encompasses a range of innovations, i.e., telemedicine, electronic health records (EHRs), health-related apps, wearable devices, and health information systems, enabling more personalized and real-time health monitoring. By leveraging advanced data analytics and ML methods, digital healthcare aims to diagnose and prevent diseases, optimize treatment strategies, and improve overall quality of the healthcare sector. Nonetheless, its widespread adoption encounters challenges, particularly concerning medical data management and privacy issues. 

FL allows healthcare organizations to collaboratively train high-quality medical ML models without accessing private medical data \cite{antunes2022federated}. Despite this, the varying performance of FL models across patients with diverse ethnic or racial backgrounds introduces disparities in the quality of treatment recommendations, leading to serious social inequality challenges. The article in \cite{zhang2024unified} developed a unified framework to enhance fairness at multiple levels of FL training, i.e., client-level, attribute-level, and multilevel fairness. The research proposed unified objective algorithm, integrating diverse fairness levels within FL environment. The algorithm enhanced overall efficacy by utilizing distributionally robust optimization and a unified uncertainty set while maintaining accuracy levels and consistent performance across all subgroups. The empirical results illustrated its effectiveness in mitigating biases and maintaining model performance in comparison to the standard FL benchmark FedAvg. 

\subsection{Fairness in Industrial Internet of Things}
\label{SubSec: Fairness in IIoT}
 The IoT paradigm serves as the interconnected network of physical devices embedded with sensors and actuators to gather and exchange data over the Internet. The proliferation of IoT has revolutionized various sectors, including but not limited to, education, healthcare, transportation, agriculture, marine, and retail. When applied to industrial environment, IoT becomes Industrial IoT, which refers to the integration of the cutting edge technologies, i.e., advanced sensors, AI algorithms, and real-time data analytics into industrial processes to enhance operational efficiency, reduce downtime, and increase productivity \cite{SunHWWYSD24}. Nevertheless, the conventional AI models utilized in IIoT settings generally necessitate centralized processing of data while storing and communicating to the cloud and end devices, thereby, raising significant data privacy concerns.
 
 FL in the context of IIoT offers a solution to privacy and data security challenges by (a) enabling decentralized training of ML models across multiple devices and (b) sharing only the encrypted notifications and parameters to the central server \cite{IIoT2022fusion}. It ensures data privacy, reduces bandwidth usage, and enhances scalability to uphold predictive maintenance, quality control, and energy optimization. Despite its various advantages, FL in IIoT encounters heterogeneity and adversarial attack challenges due to decentralized nature of devices involved in cross-device federated learning (CDFL), which could result in unfairly trained low-quality FL models. A blockchain-based algorithm was designed to detect the model poisoning attacks, ensure fair training, and maintain the reputation of participating devices for CDFL systems \cite{FairnessIIoT2021trustfed}. The presented framework assured fairness by detection and rectification of the outliers from the training distributions. It utilized the Ethereum blockchain smart contracts to stimulate the clients involved in FL training and maintain their on-chain reputation while ensuring their honest and active contributions during the model training. The model analysis over the large-scale IIoT dataset and several attack models illustrated its better performance in terms of (a) fairness, (b) outlier detection, and (c) distributed training and validation, in comparison to existing baseline approaches.

\subsection{Fairness in Intelligent Transport Systems} 
\label{SubSec: Fairness in ITS}
ITS have tremendously transformed how vehicles communicate with each other and with the environment, paving the way towards a new era of smart transportation. It encompasses a wide range of applications, i.e., traffic management, congestion control, real-time navigation, autonomous vehicles, and smart transportation systems by integration of AI algorithms and state-of-the-art communication technologies into transportation infrastructure, thereby, improving its efficiency, safety, and sustainability. However, traditional centralized data processing methods pose certain privacy risks and often prove impractical due to the enormous volume of data generated. Consequently, sharing low-quality data may reduce service reliability, degrade the driving experience, and potentially lead to traffic accidents.

FL in relation to ITS, serves as a decentralized approach to ML enabling vehicles, road side units and other infrastructure entities to collaboratively train models without sharing raw data \cite{FLinITS}. Whilst, the heterogeneity among different data sources can lead to incomplete or biased models, thereby, reducing the effectiveness and reliability of ITS applications. The research work in \cite{wang2021federated} envisaged a robust and fair FL scheme for Unmanned Aerial Vehicle (UAV) based crowdsensing. The algorithm initially exploited the edge computing-enabled 5G heterogeneous networks in the FL framework to provide proximal FL services for UAVs, thereby ensuring high data rates and low latency. Subsequently, an optimal incentive mechanism scheme was designed to encourage fair participation of UAVs under unevenly distributed information. Furthermore, the authors devised a model aggregation algorithm to tolerate Byzantine UAVs and fairly distribute long-term model profits across selected clients using contribution index measurement. The model was based on a reputation mechanism to recruit credible UAVs and prevent free-riding. The simulation results validated the effectiveness of the model in terms of robustness, communication efficiency, and user utility.  

\subsection{Fairness in Wireless Networks}
\label{SubSec: Fairness in Wireless Networks}
In the realm of networked global landscape, seamless communication constitutes the bedrock of cutting-edge innovations from immersive virtual experiences to real-time disaster management. Wireless networks leverage radio frequency signals to interlink diverse devices without the need for physical cables, playing a pivotal role in the communication revolution. Common wireless network technologies include cellular networks, bluetooth, Wi-Fi, and satellite communications. AI algorithms are adopted in wireless networks to optimize network performance and enable intelligent resource management by analyzing sheer volumes of data in real-time. Whilst, conventional ML algorithms rely on a centralized server for the training of raw data collected from multiple devices, this practice poses significant privacy concerns and results in substantial communication overhead while transmitting such data to central ML processors.

FL in wireless networks \cite{wirlessfed2020} not only ensures privacy preservation by allowing model training to occur locally on each device but also reduces the amount of data transmitted over the network, resulting in more efficient use of network resources and reduced latency. Since only a subset of devices, often those with better resources, is selected in the FL training process, resulting in biased models that do not generalize well across diverse network conditions. In the context of wireless networks, fairness in FL involves providing equal opportunities for all participating nodes to contribute to the model training process, regardless of their varying computational capabilities, network conditions, and data quality, thereby, preventing bias and ensuring that the trained models perform well across all devices and user groups. The research in \cite{zhu2022online} envisaged a technique to tackle asynchronous client selection problem, significantly straggler issue, in wireless networks while retaining long-term fairness. This technique utilized an asynchronous model aggregation technique to support wireless FL training with multiple sub-channels, where the local model updates from participants were aggregated on a first-come first-served basis. A Lyapunov optimization-based mechanism was utilized to convert the offline problem into an online optimization problem. Moreover, the client selection challenge was optimized by converting the latency minimization problem into multi-armed bandit problem and utilizing the upper confidence bound policy. The proposed technique theoretically demonstrated the achievement of sub-linear regret performance and ensured the stability of virtual queues while assuring long-term fairness and training convergence. 

The mentioned pioneering researches in cross-disciplinary domains underscore the significance of incorporating fairness considerations into FL frameworks to mitigate biases and enhance inclusivity, thereby fostering more trustworthy, fair, and robust AI systems. The concerted effort to embed fairness into FL systems remains instrumental in leveraging the full potential of state-of-the-art technologies, ultimately benefiting a diverse range of users and stakeholders.

\section{Fairness Evaluation in Federated Learning} 
\label{Sec: Fairness Evaluation Metrics}
Fairness, being a concept premised on ethics and social choice theory, remains challenging to evaluate, particularly in the context of FL systems. Delineating a robust set of performance evaluation measures is essential to ensure the validity of existing fairness-aware FL algorithms. In this section, we explore the recent trends in the development of fairness evaluation metrics, where several metrics are being adopted by different fairness-aware FL systems, each tailored to suit various scenarios and applications.

\subsection{Generic Evaluation Metrics}
\label{SubSec: Generic Evaluation Metrics}
Among the various fairness evaluation metrics in FL, \textit{Accuracy} is a widely adopted metric in the literature to assess the model performance and measure each client's contribution to the entire model \cite{huang2020efficiency,shi2023fairness,ACM2024privacy,GAOaccuracyICSOC,Accuracy_2025}. 
In addition to accuracy, \textit{Efficiency} is another frequently utilized metric for the evaluation of FL training process. It evaluates performance od training framework based on either the number of training rounds or the total time taken for the training process. Some researchers aim to reduce the training time \cite{mao2024federatededge,Liuevaluation2022}, while others prioritize to decrease the training iterations \cite{li2020fair_ICLR,huang2022stochastic,wang2021FedFV}. It is pertinent to mention that while these metrics do not directly assess fairness, they assist in achieving fair outcomes by ensuring stable and effective model training.

\subsection{Distance Metrics}
\label{SubSec: Distance Metrics}
Several distance metrics are employed to quantify the equitable distribution of model contributions and performance across FL participants. These metrics measure fairness between models by assessing the similarities/disparities in model performance parameters and evaluating the contributions of individual client. Let $\phi^* = (\phi^*_1, \phi^*_2, \phi^*_3, ....., \phi^*_n)$ and $\phi = (\phi_1, \phi_2, \phi_3, ....., \phi_n)$ be the vectors of two groups of $n$ clients, the discrepancy between these vectors can be calculated by following metrics.

\vspace{1mm}
\noindent {\em Euclidean Distance}. Euclidean distance calculates the shortest distance between two points in a multidimensional space. In the context of FL, it can be utilized to evaluate how closely aligned or disparate the data representations are across different federated nodes. In \cite{zhou2022personalized},  the Euclidean distance between the recovered and lost weight functions was calculated to evaluate the recovery efficiency within the FL system. It is calculated as Equation (\ref{Euclidean Distance Eq}):
    \begin{equation} \label{Euclidean Distance Eq}
       D_E = {\sqrt{\sum_{i=1}^n (\phi^*_i-\phi_i)^2}}
    \end{equation}

\vspace{1mm}
\noindent {\em Manhattan Distance}. Manhattan distance measures the distance between two vectors by calculating the sum of absolute differences of their corresponding elements. The work in \cite{Huang_2023_ICCV} utilized the Manhattan distance metric in FL settings in a bid to characterize gradients by capturing differences in high-dimensional spaces. Manhattan distance can be calculated by using Equation (\ref{Manhattan Distance Eq}): 
    \begin{equation} \label{Manhattan Distance Eq}
       D_M = {\sum_{i=1}^n |\phi^*_i-\phi_i|}
    \end{equation}
    
\vspace{1mm}
\noindent {\em Cosine Similarity/Distance}. Cosine similarity, also referred to as cosine distance, determines whether two vectors point in the same direction by measuring the cosine of the angle between them. It helps to find similar items in multidimensional space based on the direction the vectors are pointing in. The research utilized cosine similarity to detect gradient conflicts to mitigate bias and promote fair aggregation of gradients among FL clients \cite{wang2021FedFV}. The formula to calculate cosine similarity can be expressed as Equation (\ref{cosine similarity Eq}):
\begin{equation} \label{cosine similarity Eq}
    D_C = {1-cos(\phi^*,\phi)} = \frac{\sum_{i=1}^n (\phi^*_i.\phi_i)}{\sqrt{\sum_{i=1}^n(\phi^*_i)^2}.\sqrt{\sum_{i=1}^n\phi_i^2}}
    \end{equation}
    
\vspace{1mm}
\noindent {\em Maximum Difference}. Maximum difference formula calculates the maximum distance between the targeted value $\phi_i^*$ and the actual value $\phi_i$ for each $i$ in a set of elements $n$, expressed in Equation (\ref{Max. diff Eq}). The work in \cite{Liuevaluation2022}, utilized max. difference metric to evaluate the disparity between clients by measuring how each client's contribution deviates from an ideal target. A smaller value indicates a more equitable distribution of model performance across clients. 
    \begin{equation} \label{Max. diff Eq}
      D_{Max} = \max_{i=1}^n|\phi_i^*-\phi_i|    
    \end{equation} 

\subsection{Average Variance} 
\label{SubSec: Average Variance}
Average Variance (AV) evaluates the variability of a set of values over time $t$ to quantify how much individual data points differ from the overall average. 
In the context of fairness in FL, it typically quantifies the level of fairness during model optimization, where a given model is deemed fairer than others if its variance is less than others. The work in \cite{wang2021FedFV} adopted average and variance to assess the performance disparities across various FL participants. The formula for measuring AV is depicted in Equation (\ref{Average Variance Eq}):     
\begin{equation} \label{Average Variance Eq}
    AV = \frac{1}{n} \sum_{i=1}^{n} (F_{i}(t) - \bar{F}(t))^2 
\end{equation}
where $F(t)$ represents the accuracy of a model on client $i$ at time $t$, and $\bar{F}(t)$ is the average accuracy across all $n$ clients.

\subsection{Statistical Parity Difference}
\label{subSec: Statistical Parity Difference}
Statistical Parity Difference (SPD), also referred to as Demographic parity, quantifies the difference in positive outcomes across various demographic groups and can be calculated as Equation (\ref{Statistical Parity Difference Eq}):
\begin{equation}  \label{Statistical Parity Difference Eq}
    SPD = |P(\hat{Y} = 1|A = 0) - P(\hat{Y} = 1|A = 1)|
\end{equation}
herein $P(\hat{Y} = 1|A = 0)$ refers to the probability of a positive outcome for the unprivileged group, and $P(\hat{Y} = 1|A = 1)$ denotes the probability of a positive outcome for the privileged group. 

\subsection{Equal Opportunity Difference} 
\label{SubSec: Equal Opportunity Difference}
Equal Opportunity Difference (EOD) assesses the performance of a binary predictor $\hat{Y}$ concerning the sensitive attribute $A$ and the actual outcome $Y$. A predictor is supposed fair if the true positive rate is not influenced by the sensitive attribute $A$. It can be computed by Equation (\ref{EoD Eq}):
\begin{equation} \label{EoD Eq}
    EOD = P(\hat{Y} = 1|A = 0,Y=1) - P(\hat{Y} = 1|A = 1, Y=1)
\end{equation}
here $P(\hat{Y} = 1|A = 0,Y=1)$ denotes the probability of a positive outcome for unprivileged group, and $P(\hat{Y} = 1|A = 1, Y=1)$ indicates the probability of positive outcome for a privileged group.  

The work in \cite{ezzeldin, hamman2024demystifying_ICLR24} utilized SPD and EOD as fairness evaluation metrics to measure the performance of the envisaged fair client selection algorithm and demonstrate its effectiveness in addressing bias in FL scenarios. The values nearer to zero suggest greater fairness and positive values in these metrics indicate that the unprivileged group outperforms the privileged group.

\subsection{Pearson Correlation Coefficient} 
\label{SubSec: Pearson Correlation Coefficient}
The Pearson Correlation Coefficient (PCC) measures the strength and direction of the linear relationship between two variables. In the context of SV and contribution assessment methods, PCC calculates how closely the predicted contribution values align with the true SVs. 
In \cite{shi2022fedfaim}, the authors applied PCC with true SV as an assessment metric in a bid to measure the performance of contribution evaluation methods. A higher PCC value indicates greater fairness in assessing the client's contribution to the FL model, formulated as Equation (\ref{Pearson Correlation Coefficient Eq}):
\begin{equation} \label{Pearson Correlation Coefficient Eq}
       PCC = \frac{\sum_i (\phi^*_i-\overline{\phi^*})(\phi_i-\overline\phi)}{S_{\phi^*_i} \times S_{\phi_i}}    
\end{equation} 
In above equation, $\phi_i$ represents the SV for client $i$, $\phi^*_i$ shows the ground-truth SV, 
$\overline{\phi^*}$ and $\overline{\phi}$ refer to the mean values of $\phi^*$ and $\phi$, respectively. Moreover, $S_{{\phi^*}_i}$ and $S_{\phi_i}$ denote relative standard deviations.

\subsection{Jain’s Fairness Index}
\label{SubSec: Jain’s Fairness Index}
Jain’s Fairness Index (JFI) metric evaluates fairness in resource allocation among $n$ participants. The work in \cite{shi2023fairness} adopted JFI to assess the level of selection fairness gained by each approach after model convergence, in a bid to examine the level of fairness, computed as Equation (\ref{Jain’s Fairness Index Eq}):
\begin{equation} \label{Jain’s Fairness Index Eq}
    JFI = \frac{(\sum_{i=1}^n F_i (t))^2}{n.\sum_{i=1}^n (F_i (t))^2} 
\end{equation} 
where $F_i (t)$ represents local objective function of client $i$. The JFI values range from $0$ indicating highly unfair to $1$ representing most fair.

\section{Open Research Directions} 
\label{Sec: Open Research Directions}
\begin{figure}[!bt]
\centering
    \includegraphics[width=1\textwidth, trim = 2cm 3cm 2cm 4cm,]{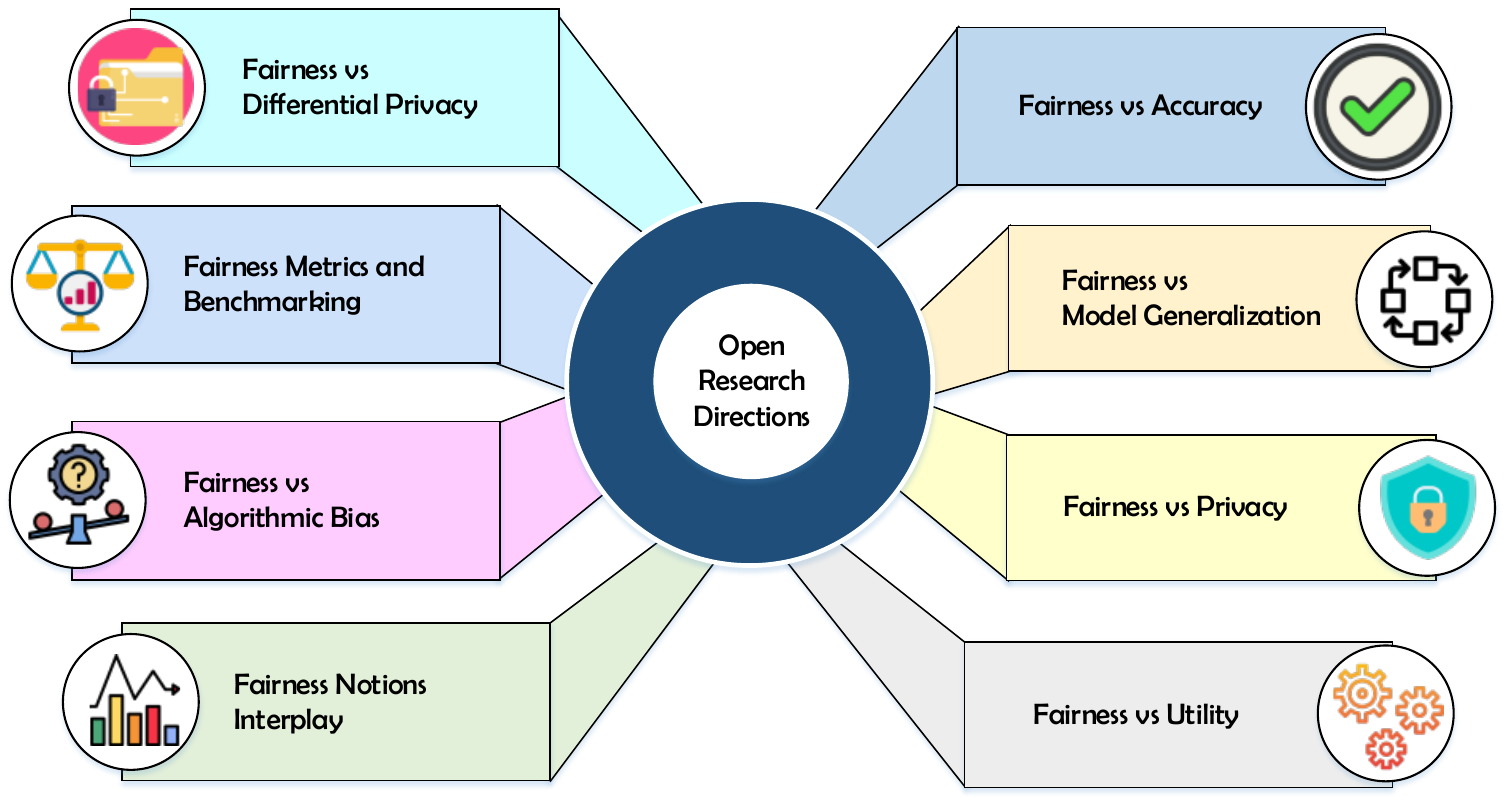}
    \caption{Open research directions for fairness in federated learning.}
    \label{Open research directions}
\end{figure}

Despite extensive research conducted in recent years, as thoroughly discussed in Section \ref{Sec: FAFL}, fairness-based client selection in an FL environment is still in its nascent stages, leaving
several open challenges. As FL continues to evolve, ensuring fairness across diverse clients considering key parameters, i.e., {\em accuracy, privacy, model generalization,  utility, differential privacy, evaluation metrics and benchmarking, model performance disparities, and the interplay of fairness notions} remains a complex challenge, which requires further exploration. 
In this section, we elaborate on the crucial open research directions in FL, depicted in \textbf{Figure \ref{Open research directions}}, to explore how each challenge interrelates with fairness and the potential trade-offs involved. By discussing the mentioned aspects in detail, we provide insights into the state-of-the-art research landscape and identify prospective future research areas in a bid to create robust, equitable, and high-performance FL models.

\subsection{Balancing Fairness and Accuracy}
\label{SubSec: Balancing Fairness and Accuracy}
A fair but less-performing model is not ideal, emphasizing the significance of balancing the trade-offs between accuracy and fairness in the FL environment. Ensuring fairness tends to maintain equitable outcomes across participants especially those with less representative data or limited resources. However, improving fairness may compromise the model accuracy, presenting an open research challenge. The research in \cite{lewis2024ensuring} highlights the interrelation between fairness and model accuracy, demonstrating a decline in overall model accuracy while improving fairness and vice versa.

The work in \cite{Changjian} strives to maintain a balance between accuracy and fairness on multiple clients' subgroups by introducing a bi-level optimization algorithm-based fair predictor. The lower-level subgroup-specific predictors are trained on limited data, while the upper-level fair predictor is adjusted to align with all subgroup-specific predictors. Whilst the empirical evaluations indicate the improvement in fair predictor without sacrificing accuracy, the method is bounded by certain conditions, i.e., the assumption of similar ground truth predictors (A-Bayes predictors) across different subgroups that could lead to non-trivial scenarios upon violation. This, in turn, presents a gap for the development of adaptive strategies required to maintain fairness and achieve competitive model performance.
 
\subsection{Fairness and Privacy Trade-offs}
\label{SubSec: Fairness and Privacy Tradeoffs}
Ensuring fairness in FL can heighten privacy risks, as it frequently requires collecting clients' sensitive demographic information that may be extraneous to the related task. This gathered information is utilized to navigate model adjustments and reduce bias, nevertheless, increasing the potential for privacy breaches or exposure to sensitive data. The work in \cite{chang2021privacy} indicates that fair models increase privacy risks for underprivileged subgroups.

Fairness-aware models necessitate consistent performance across all subgroups, however, limited data for underprivileged groups can result in overfitting to the training data of privileged groups, thereby raising privacy concerns. \cite{esipova2023disparate} proposes a solution to this issue by considering cross-model fairness (where the cost of integrating privacy to a non-private model should be equitably distributed among different groups).
They explore gradient misalignment as a key factor in disparate impact within differentially private stochastic gradient descent and envisage global scaling method to mitigate it. The empirical results demonstrate that the 
proposed method improves fairness in terms of accuracy and loss parameters without requiring protected groups' data and reduces disparate impact for all groups. However, lacks in fully eliminating biases from data collection or modeling assumptions, making independent fairness validation necessary for models with global scaling to prevent unintended disparities. Ultimately, maintaining fairness across all clients while respecting individual clients' privacy remains in its infancy, entailing innovative approaches that balance these competing priorities for trustworthy and privacy-preserving FL systems.

\subsection{Navigating Trade-off between Fairness and Generalization}
\label{SubSec: Fairness and Generalization Tradeoffs}

A model achieving higher degree of fairness at the cost of generalization is not ideal 
for long-term sustainability in the FL environment. 
Model generalization refers to the ability of a model to perform well on unseen data. Maintaining fairness requires equitable model performance across diverse client data sources, which creates a conflict in maintaining broad generalization. 
For instance, achieving generalization often requires the model to learn broad patterns from diverse data. However, overemphasizing fairness may lead to overfitting, which, in turn, reduces the ability of the models to generalize well across other unseen data, hence compromising the models' broader applicability. 

The research in \cite{mohri2019agnostic} aims to achieve fairness by preventing the model from overfitting to any specific client at the expense of others. The global model is optimized for a mixed-client target distribution while ensuring the worst-performing client's loss does not increase. This approach only performs well with a small number of clients, and generalization becomes challenging as the client pool grows. The research in \cite{NEURIPS2019} envisages a solution to the above-mentioned challenge by designing an oracle-efficient algorithm for the fair empirical risk minimization task. The empirical evaluations demonstrate the effectiveness of the algorithm. Nevertheless, ensuring fairness across both new individuals and classification tasks requires a large number of samples, which can be difficult to obtain. Therefore, the challenge lies in designing algorithms that ensure fairness while preserving model generalization across heterogeneous clients in a bid to develop fair and resilient FL environments.

\subsection{Bridging Gap between Fairness and Utility}
\label{SubSec: Fairness and Utility Tradeoffs}
Balancing fairness and utility presents a critical yet challenging open research direction in FL. Utility focuses on maximizing the overall system performance, e.g., 
accuracy, efficiency, or convergence, while fairness aims to prevent biases by involving adjustment techniques to the training process that could affect model utility. Several studies have demonstrated that improving fairness can reduce utility, and vice versa. For instance, the research in \cite{Dehdashtian_2024_CVPR} delves into the inherent trade-offs between utility and fairness, providing several insights into how improving fairness may impact utility.

To address this issue, \cite{ACM2024privacy} offers a solution to achieve an optimal balance between fairness, utility, and privacy in FL systems.
This approach employs a fairness-aware optimization strategy by constraining model updates within a predefined fairness-preserving region. It utilizes confined gradient descent (CGD) to enforce a bounded fairness constraint, limiting the deviation between individual client models and the aggregated global model to prevent the system from disproportionately favoring dominant clients during model aggregation. The empirical results from this work demonstrate that CGD significantly reduces accuracy variance across participants and outperforms baseline methods, e.g., 
FedAvg \cite{mcmahan2017communication} and Ditto \cite{li2021dittofairrobustfederated} in terms of fairness. 
However, it strictly relies on certain theoretical upper bounds and convergence guarantees, which could restrict its adaptability to highly non-convex loss functions and struggle in extreme data heterogeneity scenarios, thus leading to suboptimal utility. Accordingly, the complexity of ensuring fairness while maintaining utility in highly heterogeneous environments continues to be an important open research challenge.

\subsection{Standardizing Fairness Metrics and Benchmarking for Federated Learning}
\label{SubSec: Fairness Metric and Benchmarking}

The decentralized nature of FL, coupled with the absence of standardized fairness metrics and benchmarking frameworks, further complicates the assessment of model performance. Fairness, as a multifaceted concept, encompasses various notions, making it difficult to devise a single metric that fully captures its nuances across diverse federated environments. Recent studies have highlighted the necessity for robust fairness metrics and benchmarking frameworks tailored to FL environments. For instance, \cite{quantifying_fairness} introduces a methodology for measuring fairness in FL systems, proposing four novel metrics, e.g., {\em individual fairness}, {\em incentive fairness}, {\em orchestrator fairness}, and {\em protected group fairness}. The study emphasizes the impact of statistical heterogeneity and client participation on fairness, providing a framework for practitioners to gain insights into system fairness at various levels of granularity. Another notable contribution \cite{bai2024benchmarking} presents a benchmark specifically designed to test the limits of current methods under high client heterogeneity and diverse datasets. This research introduces a novel data partitioning method that distributes domain datasets among clients, facilitating the evaluation of 14 domain generalization methods in FL contexts.

The aforementioned studies make a significant contribution to the advancement of fairness evaluation and benchmarking methodologies by emphasizing the need for standardized fairness evaluation techniques and robust benchmarking frameworks in FL settings. However, computational overhead, dynamic client participation, and scalability challenges remain open research questions. 

\subsection{Addressing Algorithmic Disparities in Federated Model Performance}

Addressing disparities in federated model performance is a critical concern in FL systems. These performance imbalances driven by variations in data distributions across clients can lead to biased predictions that unfairly favor certain client populations. The work in \cite{Algo_disp} emphasizes the importance of addressing performance imbalances and proposes a framework that jointly considers performance consistency and algorithmic fairness across different local clients. The research frames a constrained multi-objective optimization problem, aiming to train a model that meets fairness constraints across all clients while maintaining consistent performance.

Another significant work \cite{hu2024feature} addresses the challenges of divergent model updates by integrating class average feature norms into the loss function as a regularization strategy. The research provides theoretical convergence guarantees and demonstrates improved test accuracy through extensive experiments. However, the effectiveness of the algorithm is contingent on several assumptions and hyperparameters, which, if unmet, could affect fairness and model performance. These limitations emphasize the need for approaches that not only improve accuracy but also consider fairness, performance consistency across clients, and model robustness. Therefore, mitigating model performance disparities across diverse client populations remains a key challenge, making it an exciting direction for future research in FL systems.

\subsection{The Interplay Between Fairness and Differential Privacy}
\label{SubSec: Fairness and Differential Privacy}

Maintaining a balance between fairness and differential privacy (DP) in FL systems is paramount. Whilst DP plays a crucial role in safeguarding individual data, its indiscriminate noise injection can exacerbate biases, leading to unfair outcomes. Traditional DP mechanisms introduce random noise in model updates to prevent data reconstruction, which can degrade the model's performance disproportionately across clients, particularly affecting those with underrepresented or non-dominant data distributions. For instance, \cite{DP_paper_2025} investigates the complexities involved in balancing these objectives in FL. The study mathematically analyzes how privacy perturbations, i.e., DP noise, impact model fairness. It reveals that stronger DP protection degrades fairness by disproportionately affecting model utility for certain clients.

The research in \cite{Diff_pri} tries to maintain a balance between fairness and DP by proposing a robust clustering-based algorithm that clusters clients based on their model updates and training loss values. It reduces the server’s uncertainties in clustering clients’ model updates by employing larger batch sizes with a Gaussian mixture model, to mitigate the impact of DP noise and potential clustering errors. Theoretical analyses and extensive experiments on benchmark datasets demonstrate that proposed model reduces disparities while maintaining privacy guarantees with minimal computational overhead. Nevertheless, it relies on assumption that the number of clusters is known in advance, which may not always be practical in real-world FL settings where client distributions are dynamic. Furthermore, despite achieving significant improvements in fairness, some accuracy disparities between majority and minority groups persist, particularly in high-noise settings with strong privacy guarantees. Therefore, future work requires extensive exploration to develop robust methods that effectively harmonize DP and fairness in FL scenarios, particularly, by incorporating adaptive DP mechanisms.

\subsection{Blend of Synergy and Conflict Among Fairness Notions}
\label{SubSec: Trade-offs among Different Fairness Notions}
Navigating the interplay of various fairness notions within the context of FL remains a significantly unexplored area of research. From one perspective, distinct fairness notions can complement each other by promoting equitable outcomes across various levels. These notions, when aligned, can work synergistically to ensure that all participants are treated justly and model performance benefits everyone equitably. Nevertheless, these fairness notions may come into conflict, i.e., optimizing one could inadvertently undermine others. For instance, enhancing group fairness might reduce individual fairness by compromising personalized treatment. Striving for performance distribution fairness could diminish contribution fairness, neglecting the distinct inputs of clients. This delicate balance of synergy and conflict between fairness dimensions is crucial for designing more equitable and efficient FL systems. Despite its importance, the research addressing how to manage these trade-offs effectively remains in its infancy, leaving a critical gap for future exploration.

\section{Conclusion} 
\label{Sec: Conclusion}
With the widespread adoption of federated learning (FL) in cutting-edge technologies, ensuring fairness has become a critical concern across diverse client populations. This survey article delineates the multifaceted dimensions of fairness in FL ranging from theoretical notions to practical implementations. We provide a comprehensive exploration of fundamental sources of bias and their impact on the model performance. Moreover, we summarize and categorically segregate state-of-the-art fairness-aware strategies based on techniques utilized, examine issues addressed, articulate outcomes achieved, and explore their respective limitations in a bid to offer a detailed understanding of implications pertinent to fairness in FL. We discuss the evaluation metrics extensively utilized in literature to assess the performance of fairness-aware algorithms, aiming to enhance the robustness and sustainability of the FL environment. Finally, we identify several key areas for future research,  paving the way for researchers to shape a fairer and more equitable future for FL.

\section*{Acknowledgments}
The authors sincerely acknowledge the generous support of Macquarie University for funding the research via its `International Research Excellence Award (Allocation No. 20235578)'. Quan Z. Sheng's work was partially supported by the Australian Research Council (ARC) Discovery Project DP230100233.  

\section*{Conflict of Interest}

The authors declare no conflict of interest.

\begin{figure}
  \includegraphics[width=35mm,height=50mm]{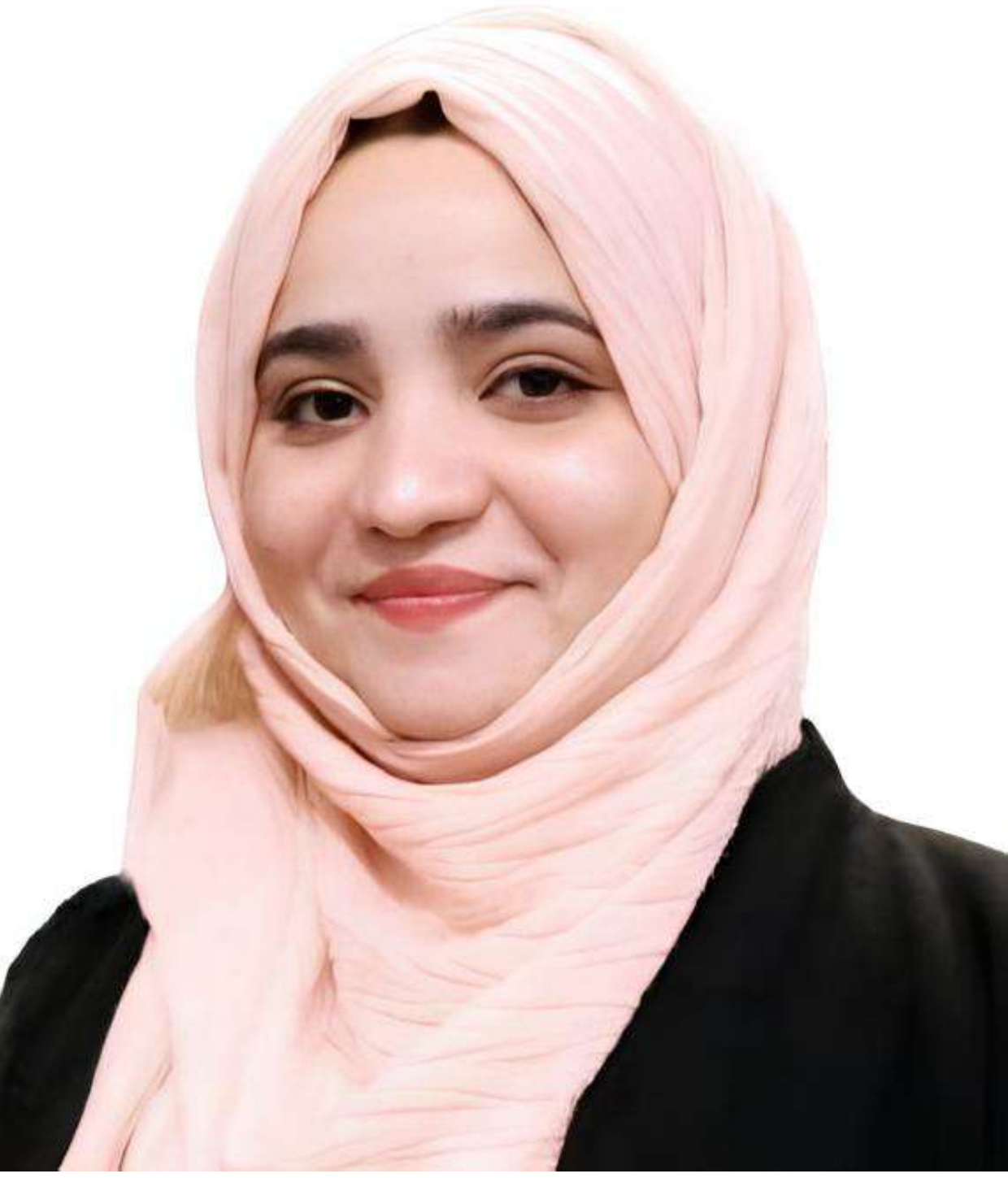}
  \caption*{Noorain Mukhtiar received a B.E. degree in Electronic Engineering and an M.E. degree in Electronic System Engineering from Mehran University of Engineering and Technology, Pakistan, in 2018 and 2022 respectively. Currently, she is pursuing her Ph.D. at the School of Computing, Macquarie University, Australia. Her research interests include, but are not limited to Federated Learning, Internet of Things, and Intelligent Transportation Systems.}
  
\end{figure}

\begin{figure}
  \includegraphics[width=35mm,height=50mm]{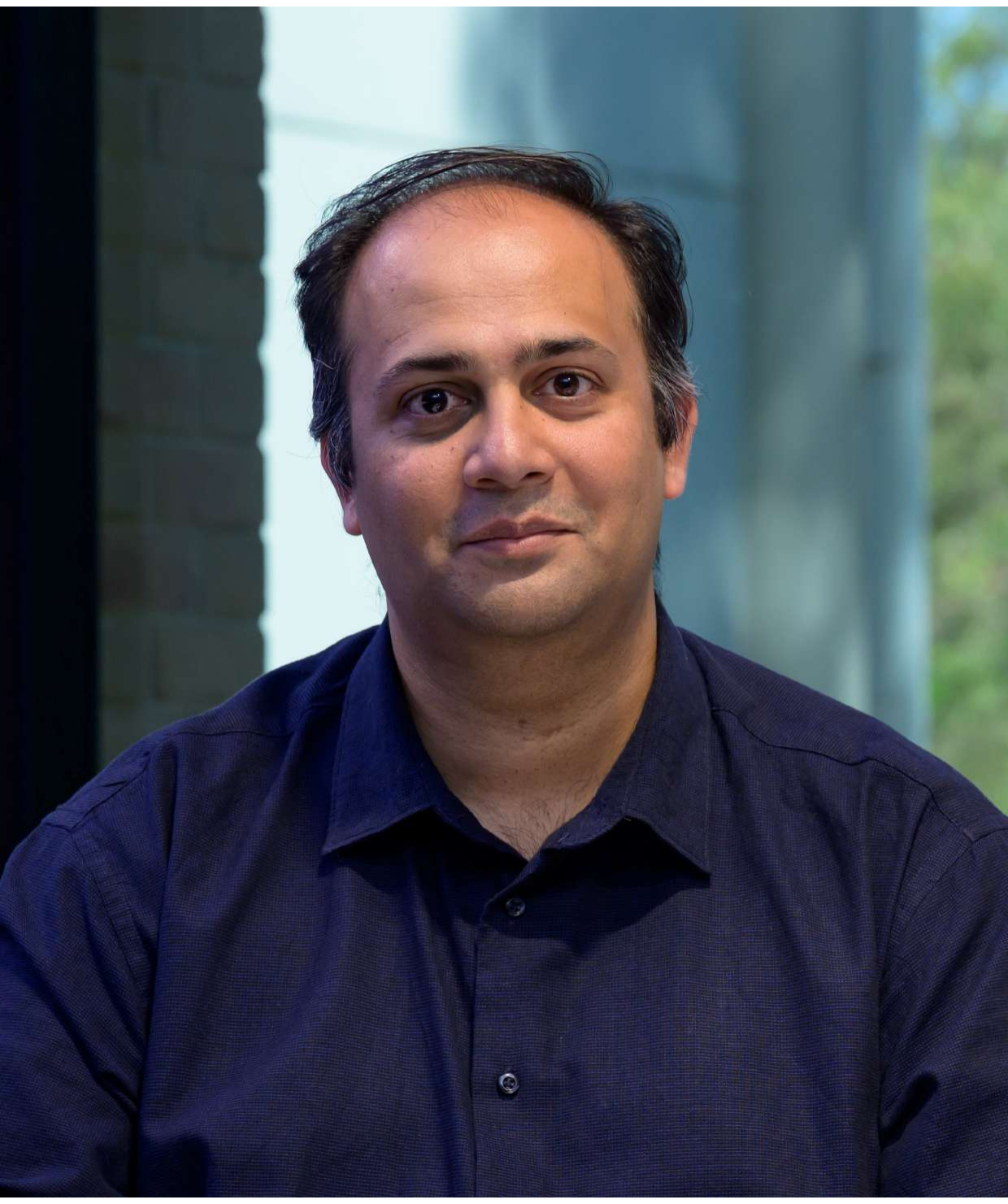}
  \caption*{Adnan Mahmood is a Lecturer in Computing – IoT and Networking at the School of Computing, Macquarie University, Sydney, Australia. Adnan’s research interests include the Internet of Things, Internet of Vehicles, Trust Management, Software Defined Networking, and Next Generation Heterogeneous Wireless Networks, among other topics. His extensive publication list includes refereed book chapters; journal articles published in prestigious venues, including but not limited to, ACM Computing Surveys, IEEE Transactions on Knowledge and Data Engineering, IEEE Transactions on Intelligent Transportation Systems, IEEE Transactions on Network and Service Management, ACM Transactions on Sensor Networks, ACM Transactions on Cyber-Physical Systems; and conference papers.}
\end{figure}

\begin{figure}
  \includegraphics[width=35mm,height=50mm]{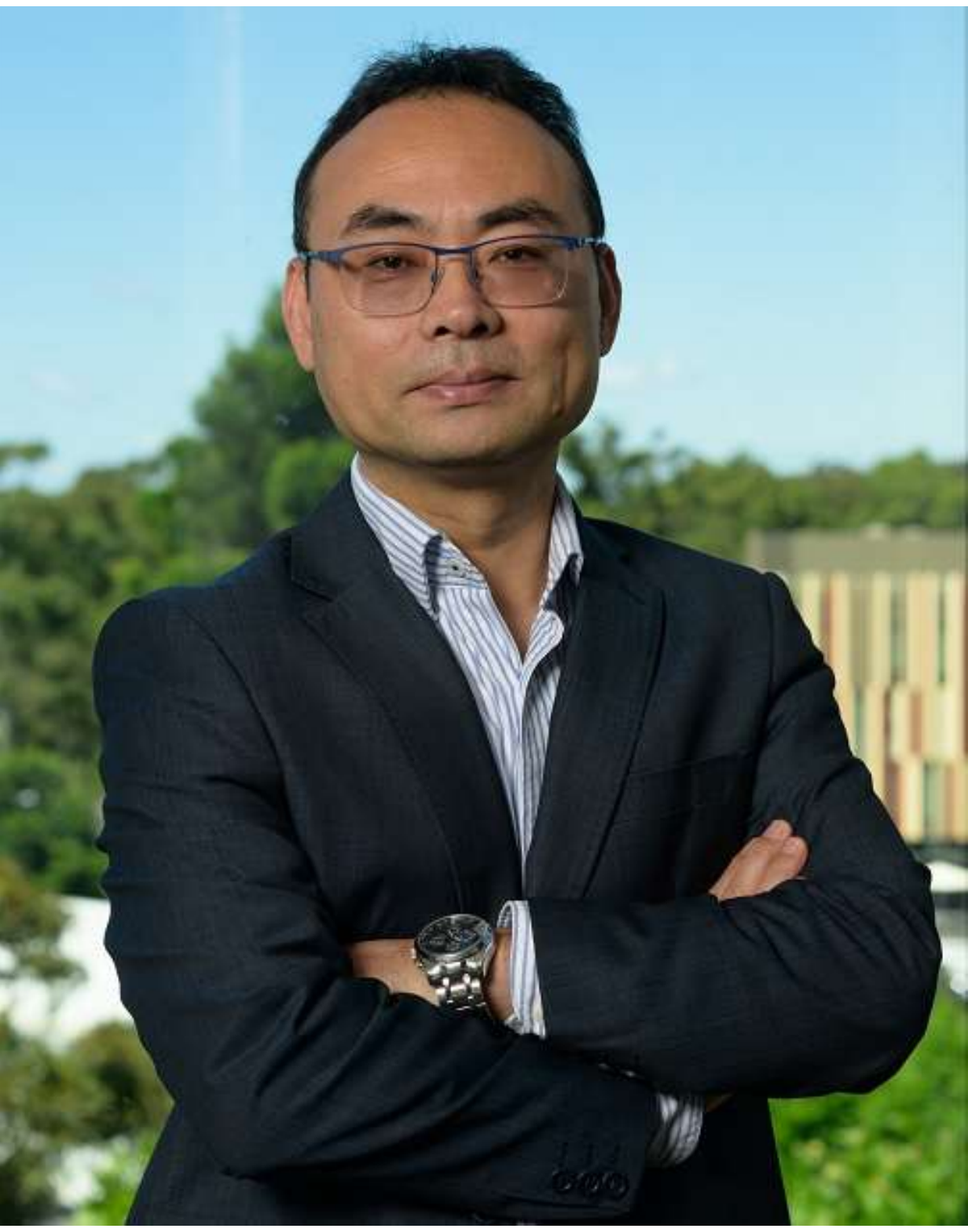}
  \caption*{Quan Z. Sheng is a Distinguished Professor and Head of School of Computing at Macquarie University, Australia. His research interests include Service Computing, Web Technologies, Machine Learning, and Internet of Things. Michael holds a PhD degree in computer science from the University of New South Wales. Distinguished Prof Sheng is the recipient of ARC (Australian Research Council) Future Fellowship, Chris Wallace Award for Outstanding Research Contribution, and Microsoft Research Fellowship. He is ranked by Microsoft Academic as one of the Most Impactful Authors in Services Computing and by ScholarGPS as one of the Highly Ranked Scholars.} 
\end{figure}

\end{document}